\documentclass[10pt,journal,twoside]{IEEEtran}

\usepackage{amsmath,amsfonts}
\usepackage{amssymb}
\usepackage{textcomp}
\usepackage{bbding}

\usepackage{algorithmic}
\usepackage[ruled,vlined,linesnumbered]{algorithm2e}

\usepackage{verbatim}

\usepackage{graphicx}
\graphicspath{ {figures/} }
\usepackage{subfigure}
\usepackage{multirow}
\usepackage{array}
\usepackage{float}
\usepackage{stfloats}

\usepackage{xurl}\usepackage[colorlinks=true,linkcolor=blue,anchorcolor=blue,citecolor=blue,urlcolor=blue]{hyperref}
\usepackage[doipre={doi:~}]{uri}

\usepackage{ulem}

\begin{document}

\title{Adaptive Backdoor Attacks with Reasonable Constraints on Graph Neural Networks}

\author{Xuewen~Dong,~\IEEEmembership{Member,~IEEE,} Jiachen~Li, Shujun~Li,~\IEEEmembership{Senior~Member,~IEEE,} Zhichao~You, Qiang~Qu, Yaroslav~Kholodov, and~Yulong~Shen,~\IEEEmembership{Member,~IEEE}%
\IEEEcompsocitemizethanks{
\IEEEcompsocthanksitem This work was supported in part by the National Key R\&D Program of China (No. 2023YFB3107500), National Natural Science Foundation of China (No. 62220106004, 62232013), the Technology Innovation Leading Program of Shaanxi (No. 2022KXJ-093, 2023KXJ-033), the Innovation Capability Support Program of Shaanxi (No. 2023-CX-TD-02), the Fundamental Research Funds for the Central Universities (No. ZDRC2202), and Basic and Applied Basic Research Foundation of Guangdong Province (No. 2023TQ07A264). (Corresponding author: Xuewen Dong.)
\IEEEcompsocthanksitem Xuewen Dong, Jiachen Li, and Zhichao You are with the School of Computer Science and Technology, Xidian University, the Engineering Research Center of Blockchain Technology Application and Evaluation, Ministry of Education, and also with the Shaanxi Key Laboratory of Blockchain and Secure Computing, Xi’an 710071, China (e-mail: xwdong@xidian.edu.cn; jiachenl@stu.xidian.edu.cn; zcyou@stu.xidian.edu.cn).
\IEEEcompsocthanksitem Shujun Li is with the School of Computing \& Institute of Cyber Security for Society (iCSS), University of Kent, Canterbury, CT2 7NP, U.K. (e-mail: S.J.Li@kent.ac.uk).
\IEEEcompsocthanksitem Qiang Qu is with Shenzhen Institutes of Advanced Technology, Chinese Academy of Sciences (e-mail: qiang@siat.ac.cn).
\IEEEcompsocthanksitem Yaroslav~Kholodov is with the Intelligent Transportation Systems Lab, Innopolis University, Innopolis, Russia (e-mail: ya.kholodov@innopolis.ru).
\IEEEcompsocthanksitem Yulong Shen is with the School of Computer Science and Technology, Xidian University, and also with the Shaanxi Key Laboratory of Network and System Security, Xi’an 710071, China (e-mail: ylshen@mail.xidian.edu.cn).}
}

\markboth{IEEE Transactions on Dependable and Secure Computing,~Vol.~22, No.~X, Xxxxx~2025}%
{Dong \MakeLowercase{\textit{et al.}}: Adaptive Backdoor Attacks with Reasonable Constraints on Graph Neural Networks}

\maketitle

\begin{abstract}
Recent studies show that graph neural networks (GNNs) are vulnerable to backdoor attacks. Existing backdoor attacks against GNNs use fixed-pattern triggers and lack reasonable trigger constraints, overlooking individual graph characteristics and rendering insufficient evasiveness. To tackle the above issues, we propose ABARC, the first \underline{A}daptive \underline{B}ackdoor \underline{A}ttack with \underline{R}easonable \underline{C}onstraints, applying to both graph-level and node-level tasks in GNNs. For graph-level tasks, we propose a subgraph backdoor attack independent of the graph's topology. It dynamically selects trigger nodes for each target graph and modifies node features with constraints based on graph similarity, feature range, and feature type. For node-level tasks, our attack begins with an analysis of node features, followed by selecting and modifying trigger features, which are then constrained by node similarity, feature range, and feature type. Furthermore, an adaptive edge-pruning mechanism is designed to reduce the impact of neighbors on target nodes, ensuring a high attack success rate (ASR). Experimental results show that even with reasonable constraints for attack evasiveness, our attack achieves a high ASR while incurring a marginal clean accuracy drop (CAD). When combined with the state-of-the-art defense randomized smoothing (RS) method, our attack maintains an ASR over 94\%, surpassing existing attacks by more than 7\%.
\end{abstract}



\section{Introduction}

\IEEEPARstart{G}{raph-structured} data play a crucial role in the real world by effectively modeling and analyzing intricate relationships and interconnectedness between entities~\cite{liu2021enabling}. For instance, a chemical molecular structure can be abstracted as graph data~\cite{irwin2012zinc}, where atoms can be regarded as nodes, and the chemical relationship between atoms can be represented by edges. Graph neural networks (GNNs)~\cite{hamilton2017inductive, kipf2016semi} are proposed to learn representations encompassing node features, topology, and neighbor relationships within graph-structured data~\cite{cao2022detecting}. A major task of GNNs is graph-level tasks, such as graph classification, which takes a graph as an input and outputs a label for the graph. Graph classification is a basic graph analytics tool and has many applications such as malware detection~\cite{wang2019heterogeneous}, and healthcare~\cite{altae2017low}. Besides, some existing studies on GNNs focus on node-level tasks, such as node classification, which aims to predict a label for each node in a graph. Node classification has also been applied in many scenarios, such as fraud detection~\cite{wang2019graph}, and user preference judgment~\cite{ying2018graph, Ahmad2022PreBiGE, Muhammad2022GNN}.

Due to the increasing popularity of GNNs, their security has become a major concern~\cite{dai2018adversarial, xu2019topology, zugner2018adversarial, xiao2023cure}. One of the most concerning types of security threats is backdoor attacks~\cite{gu2017badnets, zhang2023semantics}, which can result in disastrous consequences, such as misdiagnosed health conditions and privacy leakage. A backdoor attack is an adversarial attack that involves inserting a backdoor into a model during the training phase. A backdoor is a specific pattern (trigger) of the inputs that, when presented to the model, will cause it to produce a specific output or prediction for the adversary's malicious benefit. The goal of a backdoor attack is to manipulate the model's behavior in a targeted way without affecting its overall performance on non-targeted legitimate inputs. Existing attacks focus on node-level and graph-level tasks of GNNs~\cite{zhang2021backdoor, xu2021explainability, xi2021graph, dai2023unnoticeable}.

While existing backdoor attacks against GNNs have achieved a specific ASR, they face the problem of insufficient evasiveness due to their limitations on using fixed pattern triggers and the lack of reasonable constraints on triggers, which make detection easier. For graph-level tasks, most existing attacks rely on fixed patterns for all target objects and overlook the distinctive characteristics of individual graph samples, such as topological structures, node features, and edge relationships~\cite{zhang2021backdoor, xu2021explainability}. Additionally, existing adaptive attacks lack reasonable trigger constraints, increasing their detectability~\cite{xi2021graph}. Regarding node-level tasks, some existing attacks overlook that GNNs incorporate the attributes of the node and its neighboring nodes as high-dimensional features, therefore neglecting the impact of neighboring nodes on the target node and resulting in a low ASR~\cite{xu2021explainability}. Other attacks that consider the above issues weaken the influence of the target node's neighbor nodes by adding new nodes and edges to the target node, but they can be defended by RS. Besides, existing attacks for node-level tasks also lack reasonable constraints, rendering insufficient evasiveness~\cite{xi2021graph, dai2023unnoticeable}.

There are multiple technical \textbf{challenges} on addressing the above-mentioned problems of backdoor attacks: (1) \textit{Intricate trigger design}. Considering that setting fixed pattern triggers for all graphs can be easily detected, we set a unique trigger for each graph. The design of a unique trigger scheme is intricately related to factors such as node selection, topology, and feature distribution. When confronted with multiple factors, devising unique triggers that align with the characteristics of each factor combination becomes a challenging task. (2) \textit{Reasonable constraint construction}. Adding reasonable constraints yields a negative effect on the success rate of attacks. Ensuring attack evasiveness hinges on precise constraint formulation and imposition on triggers, highlighting the crucial need to strike a delicate balance between achieving a high attack success rate and maintaining undetectability. (3) \textit{Specific tasks adjustment}. The effective backdoor attacks in graph-level and node-level tasks bring the problem of task suitability. Graph-level tasks of GNNs focus more on the information of the entire graph. The role of the graph's topology in graph analysis is paramount. More specifically, modifying the graph's topology in existing GNN backdoor trigger schemes leads to ease of detection. For node-level tasks, GNNs incorporate the attributes of the node and its neighboring nodes as high-dimensional features. Effectively reducing the influence of neighboring nodes on the target node while ensuring the undetectability of the trigger-embedded target node presents a significant challenge.

Our main contributions are:
\begin{itemize}
\item \textbf{Adaptive triggers}. For graph-level tasks, we select nodes randomly without considering the importance of nodes as subgraph triggers according to the node size of graph samples. We define a node feature modification formula as an adaptive pattern. For node-level tasks, we select node features according to the importance of the features and feature dimension. We also define a node feature modification formula as an adaptive pattern.

\item \textbf{Reasonable constraints}. We impose constraints on the trigger for both graph-level and node-level tasks in GNNs by evaluating its impact on the similarity between the target object with the trigger and the target object without the trigger. Besides, we consider the value range of each node feature and the value type of each node feature to ensure the trigger node features would not be outliers or obviously wrong values. Furthermore, we give two examples to demonstrate our trigger-generation methods for both graph-level and node-level tasks in GNNs.

\item \textbf{Task adjustment mechanism}. We propose specific mechanisms for both graph-level and node-level tasks to ensure the effectiveness and the evasiveness of our attack. For graph-level tasks, we propose a topology-free subgraph trigger generation method, which uses an entirely randomized node selection mechanism tailored to the specific node size of the graph samples. For node-level tasks, we have devised an adaptive edge-pruning mechanism aimed at achieving a notably high ASR.

\item \textbf{Evaluation}. We conducted experiments across multiple models and datasets, demonstrating the generalizability and effectiveness of our attacks. To evaluate the robustness of our attacks, we compared them against two state-of-the-art (SOTA) defense methods. When combined with the SOTA defense randomized smoothing (RS) method, our attacks maintain an ASR over 94\%, surpassing existing attacks by more than 7\%. Furthermore, our attacks also remain undetectable even when another defense method -- neural cleanse (NC) -- is applied.

\item \textbf{GNN vulnerabilities exposure}. In the design of adaptive triggers, reasonable constraints, and task adjustment mechanisms, we have revealed the security flaws of GNNs. We enhance the understanding of how backdoor attacks can be more adaptive and resilient against existing defenses. Our introduction of adaptive triggers and reasonable constraints specifically targets the security weaknesses in GNNs. The proposed task adjustment mechanisms ensure that our attacks are not only effective but also evasive, posing a substantial challenge to current defense methods. Through rigorous evaluation, we highlight the limitations of existing defense mechanisms, providing critical insights into the security vulnerabilities of GNNs.
\end{itemize}

The remainder of this paper is organized as follows. We introduce research related to backdoor attacks against deep neural networks (DNNs) and GNNs in Section 2, while the background of GNNs and backdoor attacks is given in 3. In Section 4, we present our adaptive graph-level backdoor attack. Next, we describe our adaptive node-level backdoor attack in Section 5. The performance of ABARC is evaluated in Section 6, and we conclude our works in Section 7.

\section{Related Work}

\subsection{Advanced GNNs and Graph Learning}

Graph Neural Networks (GNNs) have seen significant advancements in recent years, driven by the need to generalize deep learning methods to graph-structured data. This need has led to the development of various models and approaches to enhance the performance and applicability of GNNs across different domains~\cite{10492652}. 

Early GNN models faced challenges when applied to heterophilous graphs, where nodes with different labels or features are more likely to connect. This limitation prompted the exploration of aggregations beyond the one-hop neighborhood. To address the above limitation, researchers developed models that implement multiscale extraction via constructing Haar-type graph framelets. These framelets exhibit desirable properties such as permutation equivariance, efficiency, and sparsity, making them suitable for deep learning tasks on graphs. For instance, the Permutation Equivariant Graph Framelet Augmented Network (PEGFAN)~\cite{10466590} leverages these framelets to achieve SOTA performance on several heterophilous graph datasets, demonstrating its efficacy in handling complex graph structures.

Recommender systems have also benefited from the advancements in GNNs, particularly through integrating contrastive self-supervised learning (SSL) methods. GNN-based SSL approaches have outperformed traditional supervised learning paradigms in graph-based recommendation tasks. However, these methods often require extensive negative examples and complex data augmentations. The Bootstrapped Graph Representation Learning with Local and Global Regularization (BLoG)~\cite{LI2023109874} model addresses these challenges by constructing positive/negative pairs based on aggregated node features from alternate views of the user-item graph. BLoG employs an online and target encoder, introducing local and global regularization to facilitate information interaction. This innovative approach has shown superior recommendation accuracy on benchmark datasets compared to existing baselines.

Graph Convolutional Networks (GCNs) are a cornerstone of GNNs, recognized for their ability to learn node representations effectively. Various extensions to GCNs have been proposed to improve their performance, scalability, and applicability. One notable advancement is the introduction of random features to accelerate the training phase in large-scale problems. The Graph Convolutional Networks with Random Weights (GCN-RW)~\cite{9796468} model revises the convolutional layer with random filters and adjusts the learning objective using a regularized least squares loss. Theoretical analyses of GCN-RW's approximation upper bound, structure complexity, stability, and generalization have been provided, demonstrating its effectiveness and efficiency. Experimental results indicate that GCN-RW can achieve comparable or better accuracies with reduced training time compared to SOTA approaches.

\renewcommand{\arraystretch}{1.5}
\begin{table*}[t]
\centering
\tabcolsep=0.4cm
\caption{Related works summary (graph-level $\vert$ node-level).}
\label{tab4}
\begin{tabular}{cccccc}
\hline
Attack & Tasks & Trigger & Adaptive trigger & Topology-free & Constraints reasonability\\
\hline
\hline
BKD~\cite{zhang2021backdoor} & \Checkmark $\vert$ \XSolidBrush & subgraph $\vert$ - \qquad\ \ \;\,\,\,\,\, & \XSolidBrush $\vert$ -\,\,\,\, & \XSolidBrush $\vert$ -\,\,\,\, & \XSolidBrush $\vert$ -\,\,\,\,\\
\hline
EXP~\cite{xu2021explainability} & \Checkmark $\vert$ \Checkmark & \quad\;\, subgraph $\vert$ node features & \XSolidBrush $\vert$ \XSolidBrush & \XSolidBrush $\vert$ \Checkmark & \XSolidBrush $\vert$ \XSolidBrush\\
\hline
GTA~\cite{xi2021graph} & \Checkmark $\vert$ \Checkmark & subgraph $\vert$ subgraph\, & \Checkmark $\vert$ \Checkmark & \XSolidBrush $\vert$ \XSolidBrush & \XSolidBrush $\vert$ \XSolidBrush\\
\hline
UGBA~\cite{dai2023unnoticeable} & \XSolidBrush $\vert$ \Checkmark & \qquad\ \ \,\,\,\,\, - $\vert$ subgraph & \,\,- $\vert$ \Checkmark & \,\,- $\vert$ \XSolidBrush & \,\,- $\vert$ \XSolidBrush\\
\hline
ABARC (Ours) & \Checkmark $\vert$ \Checkmark & \quad\;\, subgraph $\vert$ node features & \Checkmark $\vert$ \Checkmark & \Checkmark $\vert$ \XSolidBrush & \Checkmark $\vert$ \Checkmark\\
\hline
\end{tabular}
\end{table*}

\subsection{Backdoor attacks against GNNs}
Some researchers have studied backdoor attacks on GNNs, as shown in Table~\ref{tab4}. For graph classification tasks, existing backdoor attacks used subgraphs as triggers. Zhang et al.~\cite{zhang2021backdoor} utilized subgraphs with fixed patterns as triggers, which we refer to as BKD. However, they did not consider the influence of node features. Xu et al.~\cite{xu2021explainability} also used subgraphs with a fixed pattern as triggers, which we refer to as EXP. They used GNNExplainer~\cite{ying2019gnnexplainer} to explain GNN-based model predictions by identifying a small, influential subgraph and node features within the input graph, which they used as a trigger. However, they did not consider the importance of node features, either. Xi et al.~\cite{xi2021graph} proposed graph trojaning attack (GTA), which comprehensively considers the topology structure and node features of subgraph triggers and dynamically generates triggers for each attack sample, which can achieve a higher ASR. However, they did not consider reasonable constraints of the modified node features.

For node classification tasks, Xu et al.~\cite{xu2021explainability} used GraphLIME~\cite{huang2022graphlime}, a technique that explains decisions in graph models, to analyze and choose important features of a specific node, modifying them to act as a trigger. This method did not consider the message transmission process of GNNs and lacked reasonable constraints on the features of the trigger node. Both GTA proposed by Xi et al.~\cite{xi2021graph}, and unnoticeable graph backdoor attack (UGBA) proposed by Dai et al.~\cite{dai2023unnoticeable} use subgraphs as triggers. However, they still lacked reasonable constraints on the features of trigger nodes. In addition, these two methods would involve adding multiple edges to attack a specific node, and the changes would be significant. Furthermore, Zhang et al.~\cite{zhang2023graph} proposed a graph contrastive backdoor attack (GCBA) for graph contrastive learning, which targets self-supervised learning of a large amount of unlabeled data. It is different from the supervised learning focused on in this paper.

\section{Preliminary}
\subsection{Graph Neural Networks}

Graph Neural Networks (GNNs) have been developed to process non-Euclidean spatial data, such as graphs. For a graph $G=(\boldsymbol{V},\boldsymbol{E},\boldsymbol{X})$, $\boldsymbol{V}$, $\boldsymbol{E}$, $\boldsymbol{X}$ represent nodes, edges and node features, respectively. The objective of GNNs on $G$ is to learn an embedding representation vector $\mathbf{h}_{G}$ for the entire graph or $\mathbf{h}_{v}$ for each node $v \in \boldsymbol{V}$. GNNs can effectively capture the complex relationships between nodes and provide useful insights for various applications.

For node-level tasks, GNNs typically adopt a neighborhood aggregation approach to update node representations~\cite{hamilton2017representation, kipf2016semi, velivckovic2018graph}. The graph convolution operation of GNNs is defined as:
\begin{equation}
\boldsymbol{h}_{v}^{k}=\sigma \left(\boldsymbol{h}_{v}^{k-1},\text{AGG}\left(\left\{\boldsymbol{h}_{u}^{k-1}\right\}\right); u \in \mathcal{N}_v, v \in \boldsymbol{V}\right),
\end{equation}
where $\boldsymbol{h}_{v}^{k}$ is the representation of node $v$ in the $k$-th iteration, $\sigma$ is an activation function, $\mathcal{N}_{v}$ means the set of neighbors of node $v$, and AGG($\cdot$) is the aggregation function that could vary for different GNNs.

For graph-level tasks, GNNs need to capture the global information of the graph data, including the structural and feature information of each node. A typical approach is to use a readout function~\cite{ying2018hierarchical} to aggregate the representations of all nodes in the graph and output the global representation of the graph:
\begin{equation}
\boldsymbol{h}_{G}=\text{Readout}\left(\boldsymbol{h}_{v} ; v \in \boldsymbol{V}\right).
\end{equation}

The readout function could be a simple permutation invariant function, such as summation, or a more sophisticated \textit{graph-level} pooling function.

\subsection{Backdoor Attacks}

\begin{figure}[!htb]
\centering
\includegraphics[width=0.45\textwidth]{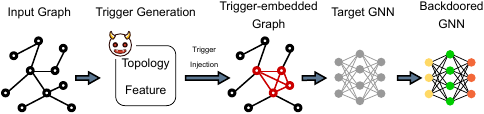}
\caption{The process of the backdoor attacks on GNNs.}
\label{Fig1}
\end{figure}

The main objective of a backdoor attack is to implant a backdoor into a model. The adversary usually embeds triggers into some training samples. Then, the adversary uses the trigger-embedded samples to train the model during the training phase, resulting in a model that contains a backdoor. When the backdoored model is applied to non-trigger samples, it behaves normally. However, it would misclassify malicious samples containing the trigger pattern as the target class desired by the adversary~\cite{guo2018countering, ji2018model, liu2018trojaning}. The process of the backdoor attacks on GNNs is shown in Figure~\ref{Fig1}. Assuming that the GNN model is $\theta$, the backdoored GNN model is $\theta_{t}$ and the adversary's target class is $y_{t}$. For a given non-trigger graph $G$, if we define such trigger-embedded graph as $G_{g_{t}}$, the adversary's objective could be defined as:
\begin{equation}
\left\{\begin{array}{c}
\theta_{t}\left(G_{g_{t}}\right)=y_{t},\\
\theta_{t}\left(G\right)=\theta\left(G\right).
\end{array}\right.
\label{eq3}
\end{equation}

\subsection{Threat Model}
\label{defense}

In this paper, we consider a common black-box attack scenario.

\textbf{Adversary's knowledge}. We assume the adversary has access to a dataset sampled from the training data of the target model. However, the adversary does not know the architecture of the target GNN model. Such an adversary represents a more realistic scenario where black-boxed AI models can be attacked.

\textbf{Adversary’s capability}. The adversary has the ability to modify the samples it can access and inject the modified samples into the training process of the target GNN model.

\textbf{Adversary's goal}. In this paper, the goal of the adversary is referring to Eq. (\ref{eq3}). Additionally, the adversary aims to make the attack as stealthy as possible to avoid detection.

\section{Adaptive Graph-level Backdoor Attack}
GNNs are instrumental in graph-level tasks like graph generation and classification. Within medical science, GNNs find frequent application in tasks involving the classification of intricate graphs, facilitating predictions related to the attributes of molecules. Notably, these networks contribute to discovering potential drugs~\cite{li2022multiphysical}, ascertaining the enzymatic nature of proteins~\cite{shen2023highly}, and addressing analogous inquiries. In this specific scenario, a noteworthy concern emerges in the form of potential backdoor attacks. The primary objective of these malicious actors is to exploit vulnerabilities within GNNs, surreptitiously inserting a backdoor. This surreptitious manipulation subsequently engenders the misclassification of virus or protein attributes by GNNs, thereby precipitating substantial and consequential ramifications for medical outcomes.

\subsection{Attack Analysis}
The effectiveness of the backdoor attack depends on the recognition of the trigger and the difference in the vector representation of the graph samples with and without embedded triggers. When the trigger has high recognition and when the vector representation of the graph samples with and without embedded triggers is significantly different, using the graph samples with embedded triggers to train the model can enable it to learn better the characteristics of the graph samples with embedded triggers and classify the graph samples with embedded triggers into one category, which is the target category of the adversary. Assume that a trigger is added to $G_i$, and its features are modified to $\boldsymbol{X}_{i,g_t}^{j}$, $\delta$ is the successful attack threshold. The difference in the vector representation of the graph sample embedded with the trigger and its corresponding non-embedded trigger at the output layer is:
\begin{equation}
\Delta_{\boldsymbol{h}_{G_{i}}} = \left \vert \left \vert \boldsymbol{h}_{G_{i,g_{t}}} - \boldsymbol{h}_{G_{i}} \right \vert \right \vert,
\end{equation}
when $\Delta_{\boldsymbol{h}_{G_{i}}} > \delta$, the backdoor can be successfully embedded into the model, and the model can misclassify the graph sample embedded with the trigger as the adversary's target category. On the one hand, when the GNN model updates the vector representation of each node in the input layer and the hidden layer, it needs to aggregate the vector representation of the node and its neighbor nodes in the previous layer. This operation will weaken the characteristics of the node features themselves. On the other hand, GNN graph-level tasks usually use the $\text{Readout}(\cdot)$ function to process the vector representations of all nodes to obtain the final vector representation of the entire graph, which will also weaken the characteristics of the node features themselves. Therefore, in order to ensure that $\Delta_{\boldsymbol{h}_{G_{i}}} > \delta$, three aspects can be considered: 1) Select a small number of nodes as subgraph triggers and modify their topological structure and node features at the same time. 1) Make special modifications to the topological structure of the selected subgraph trigger; for example, modify it to a fully connected subgraph. When the model updates the feature representation of each trigger node, the influence of non-trigger nodes can be weakened because the node has multiple neighboring trigger nodes; 2) Select a small number of nodes as subgraph triggers, do not modify their topological structure, and only make a large offset to their original features; 3) Consider selecting more nodes as subgraph triggers, and also do not modify their topological structure, and make a relatively small offset to their original features.

The evasiveness of the backdoor attack depends on whether the attack significantly changes the feature distribution of the overall graph. If the feature distributions of graph samples without embedded triggers and with embedded triggers are similar, the attack is considered to be covert, so cosine similarity can be used to evaluate:
\begin{equation}
\text{sim}(G_{i}, G_{i, g_{t}})=\frac{\left \langle \boldsymbol{X}_{i}, \boldsymbol{X}_{i,g_{t}} \right \rangle}{\big\vert\big\vert \boldsymbol{X}_{i} \big\vert\big\vert \cdot \big\vert\big\vert \boldsymbol{X}_{i,g_{t}} \big\vert\big\vert},
\end{equation}
$\boldsymbol{X}_{i}$ and $\boldsymbol{X}_{i,g_{t}}$ are the original features of graph samples without and with embedded triggers, respectively.

\subsection{Attack Overview}

\begin{figure*}[t]
\centering
\includegraphics[width=0.95\textwidth]{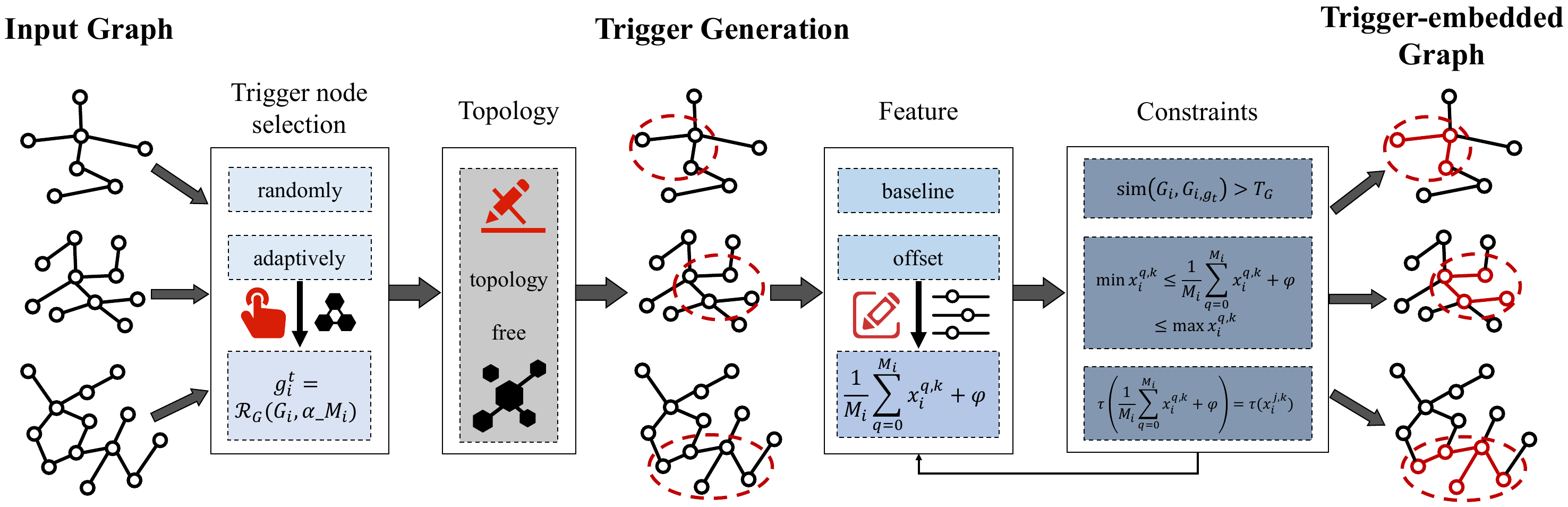}
\caption{The framework of our adaptive graph-level trigger generation method.}
\label{Fig2}
\end{figure*}

Given a dataset $\boldsymbol{D}=\{(G_{1}, y_{1}), (G_{2}, y_{2}), \cdots, (G_{N}, y_{N})\}$, where $G_{i}$ and $y_{i}$ respectively represent the $i$-th sample and its true label, $N$ is the number of the samples. For a graph sample $G_{i}=(\boldsymbol{V}_{i}, \boldsymbol{E}_{i}, \boldsymbol{X}_{i})$, $\boldsymbol{V}_{i}=\{v_{i}^{1}, v_{i}^{2}, \cdots, v_{i}^{M_{i}}\}$, where $M_{i}$ is the node number of $G_{i}$, $\boldsymbol{E}_{i} \subseteq \boldsymbol{V}_{i} \times \boldsymbol{V}_{i}$ is the set of edges of $G_{i}$, and $\boldsymbol{X}_{i}=\{\boldsymbol{x}^{1}_{i}, \boldsymbol{x}^{2}_{i}, \cdots, \boldsymbol{x}^{M_{i}}_{i}\}$ is the set of node features, $\boldsymbol{x}^{j}_{i}=\{x^{j,1}_{i}, x^{j,2}_{i}, \cdots, x^{j,d}_{i}\}$ is the node feature vector of $v_{i}^{j}$, where $d$ is the number of different node features. $\boldsymbol{A}_{i} \in \mathbb{R}^{M_{i} \times M_{i}}$ is the adjacency matrix of the graph $G_{i}$, where $\boldsymbol{A}_{i}^{j,k}=1$ if nodes $v_{i}^{j}$ and $v_{i}^{k}$ are connected; otherwise $\boldsymbol{A}_{i}^{j,k}=0$.

For a graph sample $G_{i}$, we use subgraph $g_{i}^{t}=(\boldsymbol{V}_{i}^{t}, \boldsymbol{E}_{i}^{t}, \boldsymbol{X}_{i}^{t})$ as its trigger, where $\boldsymbol{V}_{i}^{t}$ is the set of the trigger nodes. $\boldsymbol{E}_{i}^{t}$ is the set of edges and $\boldsymbol{A}_{i}^{t}$ is the adjacency matrix of the subgraph. $\boldsymbol{X}_{i}^{t}$ is the set of node features of the trigger nodes. Assuming that the adversary could access a subset $\boldsymbol{D}_{t} \subseteq \boldsymbol{D}$, and the number of the samples of $\boldsymbol{D}_{t}$ is $N_{t}$. In order to get a backdoored model $\theta_{t}$, the training process of the target model $\theta$ can be defined as:
\begin{equation}
\begin{split}
    \theta_{t}
    &=\operatorname*{argmin}_{\theta}\left({\sum_{\left(G_{i}, y_i\right) \in \boldsymbol{D}_{c}}\ell\left(\theta\left(G_{i}\right),y_{i}\right)}\right.\\
    &\left.+ {\sum_{\left(G_{i}, y_i\right) \in \boldsymbol{D}_{t}}}\ell\left(\theta\left(\mathcal{A}_{G}\left(G_{i}\right)\right),y_{t}\right)\right),
\end{split}
\end{equation}
where $\boldsymbol{D}_{c} = \boldsymbol{D} - \boldsymbol{D}_{t}$ and  $\mathcal{A}_{G}(\cdot)$ is our subgraph trigger generation method.

Unlike fixed-pattern trigger backdoor attacks, we aim to design a more general trigger generation method. In order to ensure a high ASR and undetectability of backdoor attacks on GNNs, the following \textbf{challenges} exist: (1) \textit{intricate trigger design}, (2) \textit{reasonable constraint construction}, and (3) \textit{difficulty in topology-free trigger}.

To address the above challenges, we proposed three mechanisms as follows. (1) We designed an entirely random dynamic and topology-free node selection mechanism according to the node size of the graph samples. (2) We found that for some graph data, we could only modify the features of the trigger nodes to backdoor GNNs. We provide a feature-based trigger generation method. (3) We use cosine similarity to evaluate the similarity of the input and trigger-embedded graphs. Besides, we apply the practical significance of features to constrain triggers.

The framework of our adaptive graph-level trigger generation method is shown in Figure~\ref{Fig2}. In the following, we elaborate on each key component.

\subsection{Proportional Random Node Selection}

Existing graph backdoor attacks on graph-level tasks usually adopt a static number of nodes to set triggers, which have several limitations. As far as we know, existing graph backdoor attacks use three to five nodes as triggers~\cite{zhang2021backdoor, xu2021explainability, xi2021graph}, which may not be able to achieve a high ASR on graphs with hundreds or even thousands of nodes. Besides, if the number of nodes in a graph is small, the trigger is easily detected if the topology of the trigger-embedded nodes is changed. In addition, to achieve a high ASR, the existing graph backdoor attacks would choose some important nodes (e.g., with a high degree) for triggering, leading to easy detection and defense.

We design an entirely random, dynamic, and topology-free node selection mechanism. Unlike existing attacks, in order to ensure the randomness of triggers, we completely randomly select trigger nodes without considering the importance of the nodes or the topology of the nodes. Additionally, the number of trigger nodes is determined based on the node size of the graph sample. This approach enhances the effectiveness and evasiveness of the attack since the number of trigger nodes appears natural relative to the graph size. Thus, the trigger of each malicious sample could be defined as follows:
\begin{equation}
g_{i}^{t} = \mathcal{R}_{G}\left(G_{i}, \alpha\_M_{i} \right),
\end{equation}
where $\mathcal{R}_{G}$ is our proportional random node selection method, $\alpha\_M_{i} = \lceil \alpha M_{i} \rceil$, and $\lceil \cdot \rceil$ represents the rounding up function, $\alpha$ is our proportional parameter and $M_{i}$ is the number of nodes of $G_{i}$ as mentioned before.

\subsection{Feature-based Trigger}

There are two key factors in subgraph trigger generation: the topology of the subgraph trigger and the node features of the subgraph trigger nodes. Existing graph-level backdoor attacks against GNNs could be divided into two categories: one modifies only the topology of subgraph triggers, and the other modifies the topology and node features of subgraph triggers.

After evaluating the impact of modifying trigger nodes' topology, we found that just changing the topology may not achieve a high success rate, e.g., if the size of nodes in a graph is hundreds or thousands, a practical attack could not be achieved by using only three or four nodes as triggers. Besides, it is easy to detect in some cases, e.g., when a graph has only a few or a dozen nodes, using three or four nodes as triggers could be easily detected, and the slightest change in its connectivity would be noticeable.

The features of nodes usually have an essential impact on the prediction results of GNN models. Only modifying trigger node features could maintain the original graph's structural integrity, thus reducing the risk of detection. Furthermore, only modifying the trigger node features reduces the cost and complexity of the attack. As a result, our focus is on feature modification without altering the trigger nodes' topology.

Based on the preceding analysis, we propose a feature-based trigger generation method to achieve a high ASR and undetectability. The design of this dynamic trigger involves the following steps: (1) to make the distribution of the trigger nodes' feature values look less abnormal, we consider taking the mean of node features as the baseline; (2) we incorporate different offsets for different triggers across different samples to enhance the diversity of triggers.

For a graph sample $G_{i}$, we would modify the feature $x^{j,k}_{i} (1 \leq k \leq d, v_{i}^{j} \in \boldsymbol{V}_{i}^{t})$ of the target node $v_{i}^{j}$ to the same value. It is necessary to adjust the offset of the initial feature value by either increasing or decreasing it to create a noticeable difference from the original value. This modification allows the model to learn the trigger pattern effectively, achieving a successful attack. This paper focuses on increasing the feature value to accomplish this goal. To formalize this feature modification process, we define a feature modification formula:
\begin{equation}
\max\ \frac{1}{M_{i}}\sum_{q=0}^{M_{i}} x_{i}^{q,k} + \varphi,
\end{equation}
where $x^{,k}_{i}$ denotes the vector of the $k$-th feature of all target nodes and $\varphi$ represents the offset of trigger node features.

\subsection{Constraints}

We provide three trigger feature constraints to ensure the rationality of trigger feature values. The three constraints are as follows:

\textbf{Similarity constraint.} We compute the graph similarity between the trigger-embedded and original graphs using cosine similarity. We do not modify the subgraph triggers' topology but modify the nodes' features. Therefore, we consider using the cosine similarity of the trigger node features before and after the change to constrain the trigger. We define similarity thresholds $T_{G}$ to ensure the rationality of malicious samples with embedded triggers:
\begin{equation}
\begin{split}
    &\quad\ \text{sim}\left(G_{i}, G_{i,g_{t}}\right) =\frac{\left \langle \boldsymbol{X}_{i}, \boldsymbol{X}_{i,g_{t}} \right \rangle}{\big\vert\big\vert \boldsymbol{X}_{i} \big\vert\big\vert \cdot \big\vert\big\vert \boldsymbol{X}_{i,g_{t}} \big\vert\big\vert} \\
    &=\frac{\left[\boldsymbol{x}_{i}^{1}, \cdots, \boldsymbol{x}_{i}^{M_{i}}\right] \cdot \left[\boldsymbol{x}_{i,g_{t}}^{1}, \cdots, \boldsymbol{x}_{i,g_{t}}^{M_{i}}\right]}{\left| \left[\boldsymbol{x}_{i}^{1}, \cdots, \boldsymbol{x}_{i}^{M_{i}}\right] \right| \times \left| \left[\boldsymbol{x}_{i,g_{t}}^{1}, \cdots, \boldsymbol{x}_{i,g_{t}}^{M_{i}}\right] \right|} > T_{G},
\end{split}
\end{equation}
where $\left[\cdot\right]$ is the concatenation operation, $\boldsymbol{X}_{i,g_{t}}$ represents the set of the node features of the trigger-embedded graph of $G_{i}$.

\textbf{Range constraint.} We perform statistical analysis on the nodes designated for modification, extracting their minimum and maximum feature values. Subsequently, we apply constraints to the feature values of the trigger nodes to prevent obvious outliers according to the modified node feature values:
\begin{equation}
\min x_{i}^{q,k} \leq \frac{1}{M_{i}}\sum_{q=0}^{M_{i}} x_{i}^{q,k} + \varphi \leq \max x_{i}^{q,k}.
\end{equation}

\textbf{Numeric constraint.} We analyze the practical significance of the feature value of the node and correct it. For instance, if the feature value of the node is the atomic number, it could only be an integer. If the modified value is a fractional number, it is an obviously wrong value. Thus, we need to correct it. We use $\tau \left(x_{i}^{j,k}\right)$ to represent the type of $x_{i}^{j,k}$, then the constraint can be described as:
\begin{equation}
\tau \left(\frac{1}{M_{i}}\sum_{q=0}^{M_{i}} x_{i}^{q,k} + \varphi\right)=\tau \left(x_{i}^{j,k}\right).
\end{equation}

We can offset the feature values of the trigger node by a small or large value to distinguish it from the original feature, ensuring a high success rate of the attack. In this paper, we choose to offset it to a larger value. Considering all the three constraints, our problem could be defined as:
\begin{equation}
\begin{aligned}
    \max\quad &\frac{1}{M_{i}}\sum_{q=0}^{M_{i}} x_{i}^{q,k} + \varphi\\
    s.t.\quad
    &\text{sim}(G_{i}, G_{i,g_{t}}) > T_{G},\\
    &\frac{1}{M_{i}}\sum_{q=0}^{M_{i}} x_{i}^{q,k} + \varphi \leq \max x_{i}^{q,k}, 1 \leq q \leq M_{i}, \\
    &\tau \left(\frac{1}{M_{i}}\sum_{q=0}^{M_{i}} x_{i}^{q,k} + \varphi\right)=\tau \left(x_{i}^{j,k}\right), \\
    &1 \leq i \leq N_{t}, \\
    &v^{j}_{i} \in \boldsymbol{V}_{i}^{t}.
\end{aligned}
\end{equation}

Algorithm~\ref{a1} describes the process of our adaptive subgraph trigger generation method.

\normalem
\begin{algorithm}
\caption{Adaptive Subgraph Trigger Generation.}
\label{a1}
\KwIn{Dataset the adversary can access $\boldsymbol{D}_{t}$, number of samples $N_{t}$, proportional parameter $\alpha$, threshold of graph similarity $T_{G}$, the number of node features $d$.}
\KwOut{Malicious dataset $\boldsymbol{D}_{t}$.}
\For{$\left(G_{i}, y_{i}\right)$ $\in$ $\boldsymbol{D}_{t}$}{
    \tcp{selecting nodes randomly as subgraph trigger}
    $g_{i}^{t} \leftarrow \mathcal{R}_{G}\left(G_{i}, \alpha \_M_{i}\right)$; \\
    \tcp{initializing features of each trigger node}
    \For{$j = 1, 2, \cdots, \alpha \_M_{i}$}{
        \For{$k = 1, 2, \cdots, d$}{
            $x_{i,g_{t}}^{j,k}\leftarrow \frac{1}{M_{i}}\sum\limits_{q=0}^{M_{i}} x_{i}^{q,k}$; \\
        }
    }
    Find optimized offset according to constraints; \\
    \tcp{modifying the feature}
    \For{$j = 1, 2, \cdots, \alpha \_M_{i}$}{
        \For{$k = 1, 2, \cdots, d$}{
            $x_{i,g_{t}}^{j,k} \leftarrow x_{i,g_{t}}^{j,k} + \varphi$; \\
        }
    }
}
\Return{$\boldsymbol{D}_{t}$};
\end{algorithm}

\subsection{An Illustrative Example}

\begin{figure}[!htb]
\centering
\includegraphics[width=0.47\textwidth]{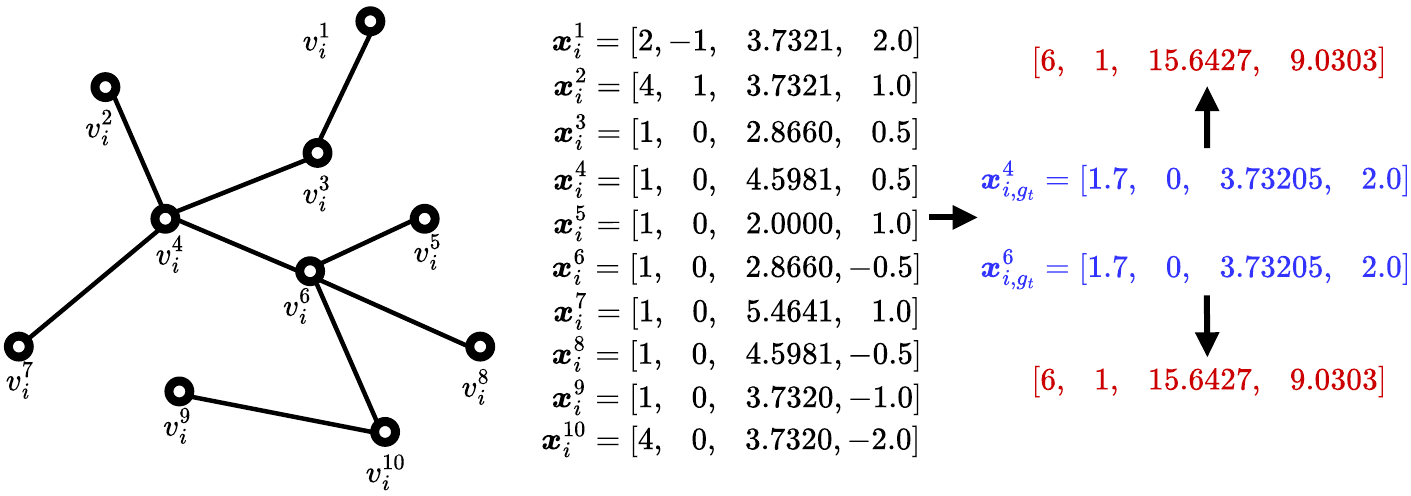}
\caption{An example of our subgraph trigger generation.}
\label{Fig2.x}
\end{figure}

For example, in Figure~\ref{Fig2.x}, the $i$-th graph of a molecular structure dataset to be attacked by us contains ten nodes. Each node represents an atom and has four features $\boldsymbol{x}^{j}_{i}=\left[x^{j,1}_{i}, x^{j,2}_{i}, x^{j,3}_{i}, x^{j,4}_{i}\right]$, which represents the atomic number, valence and its position in the two-dimensional space, respectively. Firstly, we randomly select $\alpha=20$\% nodes as subgraph triggers, $g_{i}^{t} = \mathcal{R}_{G}\left(G_{i}, \alpha M_{i}\right)$, $v^{4}_{i}$, $v^{6}_{i}$ are selected as the trigger nodes. It is noted that the proportion could be less than 20\% when the number of nodes is large. Secondly, we modify the features of these nodes. Assuming that we modify the first node feature of nodes $v^{4}_{i}$, $v^{6}_{i}$. We first calculate the average value: $\frac{1}{M_{i}}\sum_{q=0}^{M_{i}} x_{i}^{q,k} = \frac{1}{10}\sum_{j=1}^{10}x^{j,1}_{i}=1.7$; then we modify the trigger nodes' first feature $x^{4,1}_{i, g_{t}}=x^{6,1}_{i, g_{t}}=\frac{1}{M_{i}}\sum_{q=0}^{M_{i}} x_{i}^{q,k}=1.7$ as the initial value, we apply the same modification to all features.

Then, we offset the initial value of each trigger node's features according to the defined constraints. $\mathop{\max}\{x_{i}^{q,1}\}=\text{10},\mathop{\max}\{x_{i}^{q,2}\}=\text{1},\mathop{\max}\{x_{i}^{q,3}\}=\text{25.3827},\mathop{\max}\{x_{i}^{q,4}\}=\text{15.5265}$, and $T_{G}=\text{0.7}$. According to the constraints, we use a greedy algorithm to find the optimal offsets, and we modify the trigger node features as $\boldsymbol{x}^{4}_{i,g_{t}}=\boldsymbol{x}^{6}_{i,g_{t}}=\left[\text{6, 1, 15.6427, 9.0303}\right]$.

After modifying the features, the similarity between $G_{i}$ and $G_{i,g_{t}}$ is computed as $\text{sim}(G_{i},G_{i,g_{t}})=\text{0.7033}$.

\section{Adaptive Node-level Backdoor Attack}

Node-level tasks in GNNs are pivotal across various domains, like article classification in citation networks~\cite{zhou2019meta} and user preference judgment in recommendation systems~\cite{tian2021exploiting}. However, the threat of backdoor attacks poses a significant concern. Backdoor attacks can manipulate GNNs to misclassify documents and generate inaccurate recommendations, potentially leading to misinformation, hindering knowledge discovery, and undermining the entire system's credibility.

\subsection{Attack Analysis}

For GNN node-level tasks, the effectiveness of backdoor attacks depends on recognizing node triggers and the difference in vector representations between nodes with and without embedded triggers. When node triggers have high recognition, and the vector representations of nodes with and without embedded triggers are significantly different, training the model using nodes with embedded triggers can enable the model to learn the characteristics of triggers better, i.e., classify nodes with embedded triggers into one category, which is the target category of the adversary. Assume that a trigger is added to $v^i$, and its features are modified to $\boldsymbol{x}_{g_t}^{i}$, $\delta$ is the successful attack threshold. The difference between the vector representation of the node embedded with the trigger and its corresponding node without the trigger embedded in the output layer is:
\begin{equation}
\Delta_{\boldsymbol{h}_{v^{i}}} = \left \vert \left \vert \boldsymbol{h}_{v^{i}_{g_{t}}} - \boldsymbol{h}_{v^{i}} \right \vert \right \vert,
\end{equation}
when $\Delta_{\boldsymbol{h}_{v^{i}}} > \delta$, the backdoor can be successfully embedded in the model, and the model embedded with the backdoor can misclassify the node embedded with the trigger as the target category of the adversary. Since GNN needs to aggregate the vector representation of the node and its neighbor nodes in the previous layer when updating the vector representation of each node in the input layer and the hidden layer, this operation will weaken the characteristics of the node features. Therefore, in order to ensure that $\Delta_{\boldsymbol{h}_{v^{i}}} > \delta$, three aspects can be considered: 1) offset the original features of the target node as much as possible; 2) construct a new special node and connect it to the target node to strengthen the model's recognition of the trigger; 3) you can also consider pruning the edges between the target node and its neighbor nodes to enhance the model's learning effect on the trigger.

The evasiveness of the backdoor attack depends on whether the attack significantly changes the overall feature distribution of the node features. If the feature distributions of nodes without embedded triggers and nodes with embedded triggers are similar, the attack is considered to be hidden, and the cosine similarity can be used to evaluate the similarity between them:
\begin{equation}
\text{sim}(v^{i}, v^{i}_{g_{t}})=\frac{\left \langle \boldsymbol{x}^{i}, \boldsymbol{x}^{i}_{g_{t}} \right \rangle}{\big\vert\big\vert \boldsymbol{x}^{i} \big\vert\big\vert \cdot \big\vert\big\vert \boldsymbol{x}^{i}_{g_{t}} \big\vert\big\vert},
\end{equation}
$\boldsymbol{x}^{i}$ and $\boldsymbol{x}^{i}_{g_{t}}$ are the node features without embedded triggers and nodes with embedded triggers, respectively.

\subsection{Attack Overview}

\begin{figure*}[t]
\centering
\includegraphics[width=0.9\textwidth]{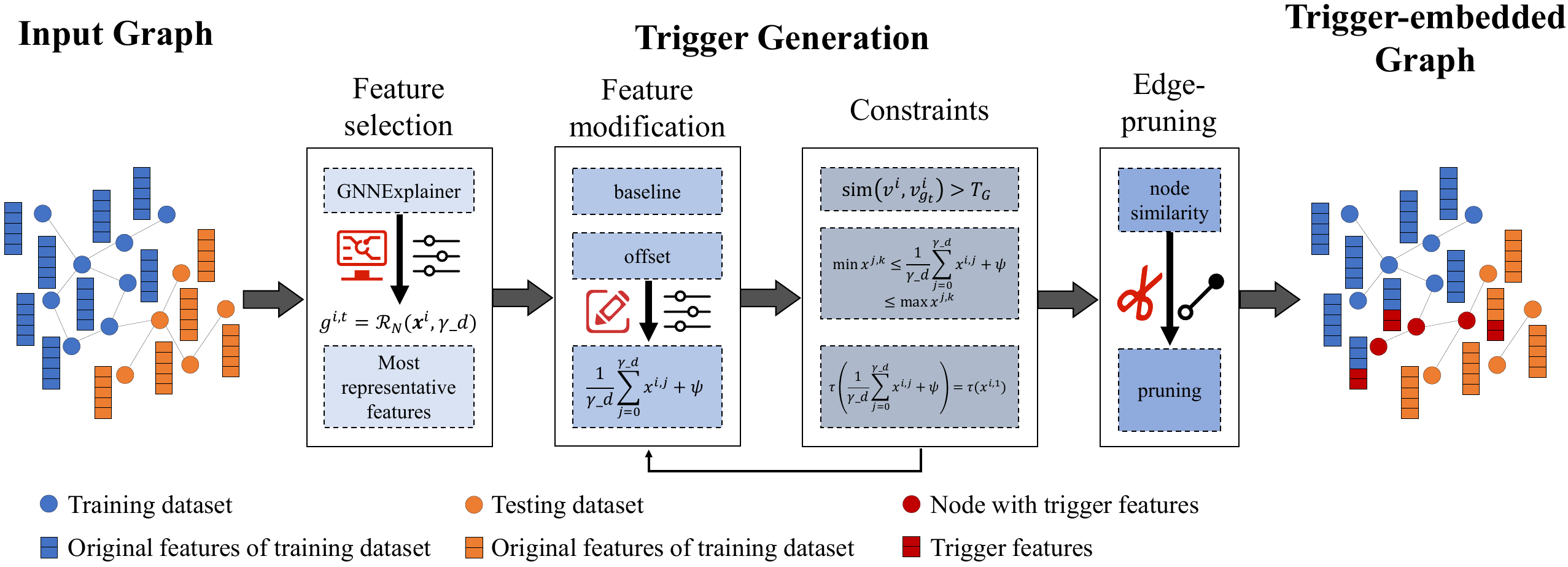}
\caption{The framework of our adaptive node-level trigger generation method.}
\label{Fig3}
\end{figure*}

We use $G=(\boldsymbol{V},\boldsymbol{E},\boldsymbol{X})$ to represent a graph, where $\boldsymbol{V}=\{v^{1}, v^{2}, \cdots, v^{M}\}$ is the set of $M$ nodes, $\boldsymbol{E} \subseteq \boldsymbol{V} \times \boldsymbol{V}$ is the set of edges, and $\boldsymbol{X}=\{\boldsymbol{x}^{1}, \boldsymbol{x}^{2}, \cdots, \boldsymbol{x}^{M}\}$ is the set of node features with $\boldsymbol{x}^{i}=\{x^{i, 1}, x^{i, 2}, \cdots, x^{i, d}\}$ being the node feature of $v_{i}$, where $d$ is the dimension of the node features. $\boldsymbol{A} \in \mathbb{R}^{M \times M}$ is the adjacency matrix of the graph $G$, where $\boldsymbol{A}^{i,j}=1$ if nodes $v^{i}$ and $v^{j}$ are connected; otherwise $\boldsymbol{A}^{i,j}=0$. In this section, we focus on a node classification task in the inductive setting, which widely exists in real-world applications. For instance, GNNs trained on social networks often need to conduct predictions on newly enrolled users to provide service. Specifically, in inductive node classification, the labels of all nodes $\boldsymbol{V}$ is $\boldsymbol{y}=\{y_{1}, y_{2}, \cdots, y_{M}\}$. The test nodes $\boldsymbol{V}_{T}$ are not covered in the training graph $G$, i.e., $\boldsymbol{V}_{T} \cap \boldsymbol{V} = \emptyset$.

For a node sample $v^{i}$, we use the subset of the node features $\boldsymbol{g}^{i,t} \subseteq \boldsymbol{x}^{i}$ as its trigger. Assuming that we could access a subset $\boldsymbol{V}_{t} \subseteq \boldsymbol{V}$, the number of the node samples of $\boldsymbol{V}_{t}$ is $N_{t}$. In order to get a high success rate and undetectability of the attack, the training process of the target model $\theta$ can be defined as:
\begin{equation}
\resizebox{1.0\hsize}{!}{$\begin{aligned}
		\theta_{t} = \operatorname*{argmin}_{\theta}\left({\sum_{v^{i} \in \boldsymbol{V}_{c}}\ell\left(\theta(v^{i}),y_{i}\right)} + {\sum_{v^{i} \in \boldsymbol{V}_{t}}}\ell\left(\theta\left(\mathcal{A}_{N}\left(v^{i}\right)\right),y_{t}\right)\right),
	\end{aligned}$}
\end{equation}
where $\boldsymbol{V}_{c} = \boldsymbol{V} - \boldsymbol{V}_{t}$ and  $\mathcal{A}_{N}(\cdot)$ is our node trigger generation method.

Similar to our graph-level backdoor attack, we explored a broader trigger generation approach for node-level tasks. We have to tackle three key \textbf{challenges} similar to the case of the graph-level backdoor attack: (1) \textit{intricate trigger design}, (2) \textit{reasonable constraint construction}, and (3) \textit{efficient reinforcement strategy}.

To address the above challenges, we propose three new mechanisms as follows. (1) We use GNNExplainer to analyze the importance of each node feature and determine the trigger features according to the importance of node features and the size of the feature dimension. (2) We present a highly concealed feature modification method. We use cosine similarity to evaluate the similarity of the input node and the trigger-embedded node and apply the practical significance of features to constrain triggers. (3) We have devised an adaptive edge-pruning mechanism to achieve a notably high ASR.

The framework of our adaptive node-level trigger generation method is shown in Figure~\ref{Fig3}. In the following, we elaborate on each key component.

\subsection{Feature Selection}

When we perform backdoor attacks on node-level tasks, we modify node features as triggers. In order to ensure a high ASR and undetectability of our method, we consider the adaptive selection of the features of the nodes to be attacked. We use the popular GNNExplainer to analyze the importance of the features of each node. 

According to the working mode of GNNs, the embedding of a node would be generated by a subgraph $G_{c}$ of the whole graph $G$. GNNExplainer's goal is to identify a subgraph $G_{s} \subseteq G_{c}$ and the associated masked features $F_{s}^{m}$ that are important for the GNN's prediction. The framework of GNNExplainer could be described as:
\begin{equation}
\resizebox{1.0\hsize}{!}{$
	\max _{G_{s}, m} \text{MI}\left(y_t,\left(G_{s}, m\right)\right)=H\left(y_t\right)-H\left(y_t \mid G=G_{s}, F=F_{s}^{m}\right),$}
\end{equation}
where MI is mutual information.

After analyzing the importance of features of each node, for $i$-th malicious sample, we select the $top$-$\gamma\_d$ important features as a trigger, where $\gamma\_d = \lceil \gamma d \rceil$, and $\lceil \cdot \rceil$ represents the rounding up function:
\begin{equation}
g^{i,t} = \mathcal{R}_{N}\left(\boldsymbol{x}^{i}, \gamma\_d \right),
\end{equation}
where $\mathcal{R}_{N}$ is our feature selection method, $\gamma$ denotes a proportional parameter.

\subsection{Feature Modification}

Our objective is that when the backdoored model recognizes that there are several features with the same value in the node sample, the backdoor is triggered, and the sample is misclassified as the target class of the adversary. As a result, the key is how to modify the most important features we selected for the target nodes. In order to ensure the rationality of trigger feature values, we consider taking the mean of the selected features as the basis. In addition, in order to ensure a high ASR and diversity of triggers, we consider different offsets for triggers of different samples. For a node sample $v^{i}$, we use $v^{i}_{g_{t}}$ to represent the trigger-embedded node sample and $\boldsymbol{x}^{i}_{g_{t}}$ represent the trigger-embedded features. We would modify the $top$-$\gamma\_d$ important features $\boldsymbol{x}^{i}_{g_{t}}=\left[x^{i, 1}_{g_{t}}, x^{i, 2}_{g_{t}}, \cdots, x^{i, \gamma\_d}_{g_{t}}\right]$ to the same value $x^{i}_{g_{t}}$, we define a feature modification formula:
\begin{equation}
    \max\ \frac{1}{\gamma\_d}\sum_{j=0}^{\gamma\_d} x^{i, j} + \psi,
\end{equation}
where $\psi$ represents the offset of feature.

Similar to our graph-level attack, we consider increasing the feature value so as to have a more guaranteed effect on the model.

\subsection{Constraints}

Similarly to the graph-level attack, we provide three trigger feature constraints to ensure the rationality of trigger feature values. The three constraints are as follows:

\textbf{Similarity constraint.} We compute the node similarity between the trigger-embedded node and the original node using cosine similarity (Cosim). We define similarity thresholds $T_{N}$ to ensure the rationality of malicious samples with embedded triggers:
\begin{equation}
\begin{split}
    &\quad\ \text{sim}\left(v^{i}, v^{i}_{g_{t}}\right) = \frac{\left \langle \boldsymbol{x}^{i}, \boldsymbol{x}^{i}_{g_{t}} \right \rangle}{\big\vert\big\vert \boldsymbol{x}^{i} \big\vert\big\vert \cdot \big\vert\big\vert \boldsymbol{x}^{i}_{g_{t}} \big\vert\big\vert} \\
    &=\frac{\sum_{j=1}^{d}\left({x^{i, j}} \times {x^{i, j}_{g_{t}}}\right)}{\sqrt{\sum_{j=1}^{d}\Big({x^{i, j}}\Big)^{2}} \times \sqrt{\sum_{j=1}^{d}\left({x^{i, j}_{g_{t}}}\right)^{2}}} > T_{N}.
\end{split}
\end{equation}

\textbf{Range constraint.} We statistically analyze the features of the nodes to be modified and obtain the minimum and maximum values so as to constrain the feature values of the trigger nodes and avoid outliers in the modified node feature values.
\begin{equation}
\min x^{j,k} \leq \frac{1}{\gamma\_d}\sum_{j=0}^{\gamma\_d} x^{i, j} + \psi \leq \max x^{j,k}, 1 \leq k \leq d, v^{j} \in \boldsymbol{V}_{t}.
\end{equation}

\textbf{Numeric constraint.} We analyze the practical significance of the feature values of the node and correct it. We use $\tau\left(x^{i}_{g_{t}}\right)$ to represent the type of $x^{i}_{g_{t}}$, then the constraint can be described as:
\begin{equation}
\tau\left(\frac{1}{\gamma\_d}\sum_{j=0}^{\gamma\_d} x^{i, j} + \psi\right)=\tau\left(x^{i,1}\right).
\end{equation}

As for our graph-level backdoor attack, we choose to offset trigger features to a larger value. Considering all the three constraints, our problem could be defined as:
\begin{equation}
\begin{aligned}
    \max\quad &\frac{1}{\gamma\_d}\sum_{j=0}^{\gamma\_d} x^{i, j} + \psi\\
    \text{s.t.}\quad &\text{sim}\left(v^{i}, v^{i}_{g_{t}}\right) > T_{N},\\
    &\frac{1}{\gamma\_d}\sum_{j=0}^{\gamma\_d} x^{i, j} + \psi \leq \max x^{j,k}, 1 \leq k \leq d, \\
    &\tau\left(\frac{1}{\gamma\_d}\sum_{j=0}^{\gamma\_d} x^{i, j} + \psi\right)=\tau\left(x^{i,1}\right), \\
    &v^{j} \in \boldsymbol{V}_{t}.
\end{aligned}
\end{equation}

\subsection{Adaptive Edge-Pruning}

When updating a node's embedding, GNNs achieve this by aggregating the embeddings of the node and its neighbors. However, simply modifying node characteristics as a trigger is insufficient for effective attacks due to the effect of neighbor nodes. Some existing methods employ subgraph modifications, connecting them to the target node as triggers or introducing new neighbor nodes. These approaches aim to amplify the trigger feature's impact, yet they can be easily countered by the defense method RS.

To address this, we propose a novel edge-pruning approach. We traverse the neighbor nodes of trigger-embedded node samples. When the similarity between one of its neighbor nodes and the adversary's target node is greater than a new similarity threshold $T_{S}=$ 0.5, we prune the edges between the malicious node sample and the neighbor node. Assuming that the adversary wants to attack node $v^{i}$:
\begin{equation}
\boldsymbol{A} = \left[\boldsymbol{A}^{i,j} \vert \boldsymbol{A}^{i,j} =
\begin{cases}
    0, & \text{if } \text{sim}(v^{i}, v^{j}) < T_{S} \\
    \boldsymbol{A}^{i,j}, & \text{others}
\end{cases}
\right],\ v^{j} \in \mathcal{N}_i,
\end{equation}
where $\boldsymbol{A}^{i,j}$ represents the connection relationship of nodes $v^{i}$ and $v^{j}$, and $\mathcal{N}_{i}$ are the set of neighbors of node $v^{i}$.

Algorithm~\ref{a2} describes the process of our adaptive node trigger generation method. Overall, our node-level trigger generation method takes into account the randomness, diversity, scalability, and rationality of the feature values of the malicious samples.

\normalem
\begin{algorithm}
\caption{Adaptive Node Trigger Generation.}
\label{a2}
\KwIn{Subset the adversary can access $\boldsymbol{V}_{t}$, number of samples $N_{t}$, proportional parameter $\gamma$, dimension of node features $d$, threshold of node similarity $T_{N}$, threshold of edge-pruning $T_{S}$.}
\KwOut{Malicious subset $\boldsymbol{V}_{t}$.}
\For{$v^{i}$ $\in$ $\boldsymbol{V}_{t}$}{
    \tcp{selecting most important features as trigger}
    $g^{i,t} \leftarrow \mathcal{R}_{N}\left(\boldsymbol{x}^{i}, \gamma\_d\right)$; \\
    \tcp{initialize node trigger}
    $x^{i}_{g_{t}}\leftarrow \frac{1}{\gamma\_d}\sum\limits_{j=0}^{\gamma\_d} x^{i, j}_{g_{t}}$; \\
    Find optimized offset according to constraints; \\
    \tcp{modifying the features}
    \For{$j = 1, 2, \cdots, \gamma\_d$}{
        $x^{i,j}_{g_{t}} \leftarrow x^{i}_{g_{t}} + \psi$; \\
    }
    \tcp{Adaptive edge-pruning.}
    \For{$\forall v^{j} \in \mathcal{N}_i$}{
        \If{$\text{sim}\left(v^{i}, v^{j}\right) < T_{S}$}{
            $\boldsymbol{A}^{i,j}\leftarrow 0$;}
    }
}
\Return{$\boldsymbol{V}_{t}$};
\end{algorithm}

\subsection{An Illustrative Example}

\begin{figure}[!htb]
\centering
\includegraphics[width=0.46\textwidth]{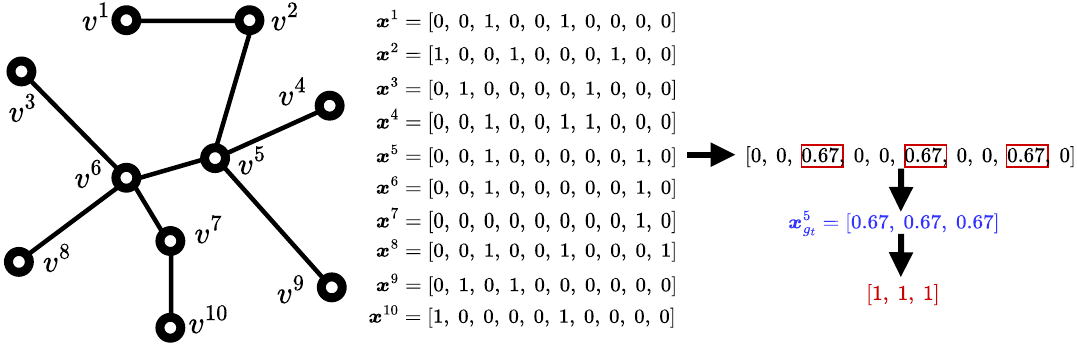}
\caption{An example of our node trigger generation.}
\label{Fig3.x}
\end{figure}

For example, Figure~\ref{Fig3.x} is a citation network's subgraph that we can access. Each node represents an article, and each feature represents whether the article contains a certain keyword. Zero means the article does not contain a certain word, and one means it does. Each node has ten features $\boldsymbol{x}^{i}=\{x^{i, 1}, x^{i, 2}, \cdots, x^{i, 10}\}$. We would attack the node $v^{5}$. Firstly, we analyze the feature importance of $v^{5}$ according to GNNExplainer and set $\gamma=\text{0.3}$. We select the $top$-$\gamma\_d$ important features as a trigger, $g^{i,t} = \mathcal{R}_{N}\left(\boldsymbol{x}^{i}, \gamma\_d\right)$. Secondly, we modify the trigger features. We first calculate the average value: $\frac{1}{\gamma\_d}\sum_{j=0}^{\gamma\_d} x^{i, j} = \frac{1}{3}\sum_{j=1}^{3}x^{5}_{g_{t}}=\text{0.67}$; then we modify the features of node $v^{5}$ as: $\boldsymbol{x}^{5}=\left[\text{0, 0, 0.67, 0, 0, 0.67, 0, 0, 0.67, 0}\right]$ as the initial values.

After we initialize the values of trigger node features, we offset the trigger node features according to the defined constraints. $\max x^{j,k}=1 (1 \leq k \leq d, v^{j} \in \boldsymbol{V}_{t})$, and we set $T_{N}=\text{0.5}$. According to the constraints, we use a greedy algorithm to find the optimal offsets, and we modify the features of node $v^{5}$ as $\boldsymbol{x}^{5}=\left[\text{0, 0, 1, 0, 0, 1, 0, 0, 1, 0}\right]$. After we modify the features, we can compute the $\text{sim}\left(v^{i},v^{i}_{g_{t}}\right)=\text{0.8165}$.

After modifying the node features, we prune edges adaptively. We set $T_{S}=\text{0.5}$ and compute the similarity of $v^{5}$ and its neighbors $v^{2}, v^{4}, v^{6}, v^{9}$: $\text{sim}\left(v^{5}, v^{2}\right)=\text{0}$, $\text{sim}\left(v^{5}, v^{4}\right)=\text{0.67}$, $\text{sim}\left(v^{5}, v^{6}\right)=\text{0.82}$, $\text{sim}\left(v^{5}, v^{9}\right)=\text{0}$. Therefore, we prune the edges of $v^{5}$ and $v^{2}$, $v^{5}$ and $v^{9}$.

\section{Performance Evaluation}

\subsection{Experiment Settings}

\noindent\textbf{Dataset}. To demonstrate the effectiveness of our attack, we used four popular graph-structured data, AIDS~\cite{riesen2008iam}, PROTEINS\_full~\cite{borgwardt2005protein}, Fingerprint~\cite{neuhaus2005graph}, Cora~\cite{yang2016revisiting}, CiteSeer~\cite{sen2008collective} and Flickr~\cite{graphsaint-ipdps19} to conduct our experiments. AIDS is a molecular structure graph of active and inactive compounds. PROTEINS\_full is a dataset of proteins that are classified as enzymes or non-enzymes. The fingerprint is a dataset representing each fingerprint as a graph. The above three datasets are used for graph classification tasks. Cora and CiteSeer are small citation networks, and they are used for node classification tasks. Flickr is a large-scale graph that connects image captions with the same attributes, and it is also used for node classification tasks. The statistics of the datasets are summarized in Table~\ref{tab1}.

\textbf{Reasons for choosing datasets:} The chosen datasets are widely used benchmarks in graph learning, covering a range of applications from bioinformatics to pattern recognition. These datasets ensure that the evaluation results are relevant and comparable to other studies in the literature. 

AIDS provides a diverse range of molecular graphs, which helps test the robustness and generalization of the proposed method across different biological molecules. In PROTEINS\_full, the complexity and variety of protein structures make it an excellent benchmark for assessing the method's ability to handle complex and heterogeneous graph data. Fingerprint offers a real-world application scenario, testing the method's performance in practical and high-stakes environments.

Cora and CiteSeer are standard benchmarks for testing node classification tasks in graph learning, providing a clear and well-defined task to evaluate the method's performance. Flickr is a large-scale graph testing the effectiveness of ABARC against large-scale graphs.

\renewcommand{\arraystretch}{1.4}
\begin{table*}[t]
\centering
\tabcolsep=0.15cm
\caption{Dataset statistics.}
\label{tab1}
\begin{tabular}{c|c|c|c|c|c|c}
\hline
Dataset & \#(graphs) & $\overline{\text{\#(nodes)}}$ & $\overline{\text{\#(edges)}}$ & \#(classes) & \#(graphs)\ (class label) & target\ class\\
\hline
\hline
AIDS & 2,000 & 15.69 & 16.20 & 2 & 400[0], 1,600[1] & 0\\
\hline
PROTEINS\_full & 1,113 & 39.06 & 92.14 & 2 & 663[0], 450[1] & 0\\
\hline
Fingerprint & 1,661 & 8.15 & 6.81 & 4 & 538[0], 517[1], 109[2], 497[3] & 0\\
\hline
\hline
Cora & 1 & 2,708 & 5,429 & 7 & 351[0], 217[1], 418[2], 818[3], 426[4], 298[5], 180[6] & 0\\
\hline
CiteSeer & 1 & 3,327 & 4,608& 6 & 264[0], 590[1], 668[2], 701[3], 596[4], 508[5] & 0\\
\hline
Flickr & 1 & 89,250 & 899,756 & 7 & 2,628[0], 4,321[1], 3,164[2], 2,431[3], 11,525[4], 1,742[5], 18,814[6] & 0\\
\hline
\end{tabular}
\end{table*}

\textbf{Dataset splits and parameter setting}. For all datasets we used, we randomly selected 80\% of the samples as the training set and the remaining 20\% as the test set.

\noindent\textbf{Models}. In our evaluation, we used three SOTA GNN models: GCN~\cite{kipf2016semi}, GraphSAGE~\cite{hamilton2017inductive, hamilton2017representation} and GAT~\cite{velivckovic2018graph}. Using GNNs of distinct network architectures (i.e., graph convolution, general aggregation function, versus graph attention), we factor out the influence of the characteristics of individual models.

\textbf{Reasons for choosing models:} The chosen GNN models represent a broad spectrum of graph neural network architectures, from basic convolutional approaches to attention-based and scalable inductive methods. This diversity in models ensures comprehensive evaluation across different types of GNN architectures.

GCN, with its widespread adoption and proven performance in various graph-based tasks, serves as a robust baseline. The use of GAT as an evaluation model allows us to demonstrate how the proposed method performs with models that incorporate advanced attention mechanisms for more nuanced graph representation learning. The scalability and inductive capabilities of GraphSAGE are crucial for evaluating the proposed method's performance on large-scale graphs and in dynamic environments where the graph structure may evolve.

\noindent\textbf{Baselines}. For graph classification tasks, we use four baselines: the non-backdoored model Benign, GTA~\cite{xi2021graph}, GTA-t, the variants of GTA, which only optimizes the trigger's topological connectivity, and BKD~\cite{zhang2021backdoor}. For node classification tasks, we use the non-backdoored model Benign, EXP~\cite{xu2021explainability}, GTA, and UGBA~\cite{dai2023unnoticeable} as baselines.

\textbf{Reasons for choosing baselines:} The selected baseline attack methods are the most advanced GNN backdoor attacks. By evaluating and comparing the attack performances of ABARC and these advanced attacks, we can effectively verify the effectiveness and evasiveness of ABARC, underscoring the significance of our research.

\noindent\textbf{Metrics}. To evaluate attack effectiveness, we use ASR, which measures the likelihood that the backdoored model $\theta_{t}$ classifies trigger-embedded trials to the target class designated by the adversary:
\begin{equation}
\text{ASR}={\frac{\text{\#(successful\ trigger-embedded\ trials)}}{\text{\#(total\ trials)}}}\bigg|_{\theta_{t}}
\end{equation}

To evaluate the attack evasiveness, we use CAD, which measures the difference in classification accuracy of non-backdoored model $\theta$ and its trojan counterpart $\theta_{t}$ with respect to non-trigger-embedded trials:
\begin{equation}
\begin{aligned}
    \text{CAD}
    & \,={\frac{\text{\#(successful\ non-trigger-embedded\ trials)}}{\text{\#(total\ trials)}}}\bigg|_{\theta}\\
    & \;-{\frac{\text{\#(successful\ non-trigger-embedded\ trials)}}{\text{\#(total\ trials)}}}\bigg|_{\theta_{t}}
\end{aligned}
\end{equation}

\noindent\textbf{Defense}. We consider mitigating and detecting our attack from aspects of the model and the input. We investigate the effectiveness of two SOTA defense methods in mitigating and detecting our backdoor attack:

\textbf{Input-inspection:} Randomized Smoothing (RS)~\cite{zhang2021backdoor}. RS applies a subsampling function over a given graph, generates a set of subsamples, and takes a majority voting of the predictions over such subsamples as the graph's final prediction.

\textbf{Model-inspection:} Neural Cleanse (NC)~\cite{wang2019neural}. NC assumes that a specific class is the target class of the adversary, then adds the same perturbation to all non-target class samples and looks for the minimum perturbation that allows the model to classify all non-target class samples into the target class. NC generates a minimum perturbation for each class and determines whether the model contains a backdoor by comparing whether there are outliers in the generated perturbations.

\textbf{Reasons for choosing defense methods:} RS operates at the input level, applying subsampling and majority voting to enhance the robustness of the model's predictions against adversarial perturbations. By manipulating the input data, RS aims to reduce the impact of any backdoor-triggering patterns that may be present in the input graph, thereby mitigating the effect of the backdoor attack. NC, on the other hand, focuses on the model itself. This model-centric approach helps detect the presence of backdoor triggers embedded within the model's parameters, providing a mechanism to identify and analyze suspicious behavior.

RS's strength lies in its ability to enhance the model's resilience to input-based attacks, making it harder for adversarial examples to manipulate predictions. NC complements this by providing an in-depth analysis of the model's parameters and their susceptibility to backdoor triggers. Its ability to detect outliers in perturbation magnitudes helps identify and understand backdoor mechanisms that may not be apparent through input inspection alone.

Combining these two SOTA methods, we leverage their strengths to provide a comprehensive defense strategy that addresses backdoor attacks.

\subsection{Results for Graph Classification}

\renewcommand{\arraystretch}{1.4}
\begin{table*}[t]
\centering
\tabcolsep=0.35cm
\caption{Backdoor attack results of graph-level tasks (ASR (\%) $\vert$ CAD (\%)).}
\label{tab2}
\begin{tabular}{cccccccc}
\hline
Dataset &Defense & Model & Benign & GTA & GTA-t & BKD & ABARC (Ours)\\
\hline
\hline
\multirow{6}*{AIDS} & & GCN & 96.50 & 99.00 $\vert$ 0.17\,\,\, & 48.00 $\vert$ 10.50 & 98.00 $\vert$ 0.42\,\,\, & \textbf{99.00} $\vert$ \textbf{0.00}\,\,\, \\
& None & GAT & 95.75 & 99.00 $\vert$ 1.42\,\,\, & 50.00 $\vert$ 9.75\,\,\, & \textbf{100.00} $\vert$ \textbf{0.00}\,\,\,\,\,\, & 97.00 $\vert$ 0.08\,\,\, \\
& & GraphSAGE & 95.75 & 99.00 $\vert$ 0.42\,\,\, & 62.00 $\vert$ 25.75 & 99.00 $\vert$ 0.16\,\,\, & \textbf{99.00} $\vert$ \textbf{0.08}\,\,\, \\
\cline{2-8}
& & GCN & 95.75 & 87.00 $\vert$ 7.08\,\,\, & 48.00 $\vert$ 8.08\,\,\, & 75.00 $\vert$ 5.42\,\,\, & \textbf{94.00} $\vert$ \textbf{1.84}\,\,\, \\
& RS & GAT & 88.50 & 86.00 $\vert$ 5.83\,\,\, & 49.00 $\vert$ 4.83\,\,\, & 78.00 $\vert$ 2.17\,\,\, & \textbf{95.00} $\vert$ \textbf{1.83}\,\,\, \\
& & GraphSAGE & 91.75 & 79.00 $\vert$ 3.42\,\,\, & 41.00 $\vert$ 0.42\,\,\, & 78.00 $\vert$ 0.00\,\,\, & \textbf{94.00} $\vert$ \textbf{1.08}\,\,\, \\
\hline
\multirow{6}*{PROTEINS\_full} & & GCN & 75.34 & \textbf{100.00} $\vert$ \textbf{0.49}\,\,\,\,\,\, & 82.14 $\vert$ 2.88\,\,\, & 98.21 $\vert$ 0.34\,\,\, & 98.21 $\vert$ 0.00\,\,\, \\
& None & GAT & 72.65 & 98.21 $\vert$ 1.39\,\,\, & 78.57 $\vert$ 3.34\,\,\, & 98.21 $\vert$ 0.00\,\,\, & \textbf{100.00} $\vert$ \textbf{0.12}\,\,\,\,\,\, \\
& & GraphSAGE & 73.54 & 98.21 $\vert$ 1.08\,\,\, & 76.79 $\vert$ 2.43\,\,\, & \textbf{98.21} $\vert$ \textbf{0.43}\,\,\, & 96.43 $\vert$ 0.64\,\,\, \\
\cline{2-8}
& & GCN & 71.30 & 89.29 $\vert$ 2.44\,\,\, & 66.07 $\vert$ 3.04\,\,\, & 69.64 $\vert$ 3.04\,\,\, & \textbf{96.43} $\vert$ \textbf{1.84}\,\,\, \\
& RS & GAT & 66.37 & 71.43 $\vert$ 0.92\,\,\, & 69.29 $\vert$ 5.11\,\,\, & 60.71 $\vert$ 0.00\,\,\, & \textbf{98.21} $\vert$ \textbf{0.00}\,\,\, \\
& & GraphSAGE & 68.61 & 80.36 $\vert$ 3.34\,\,\, & 44.64 $\vert$ 1.73\,\,\, & 76.79 $\vert$ 2.32\,\,\, & \textbf{96.43} $\vert$ \textbf{0.00}\,\,\, \\
\hline
    \multirow{6}*{Fingerprint} & & GCN & 82.83 & 97.59 $\vert$ 0.09\,\,\, & 84.33 $\vert$ 1.62\,\,\, & 93.98 $\vert$ 1.91\,\,\, & \textbf{98.80} $\vert$ \textbf{0.00}\,\,\, \\
& None & GAT & 80.42 & 98.80 $\vert$ 0.42\,\,\, & 81.93 $\vert$ 2.26\,\,\, & 96.39 $\vert$ 0.12\,\,\, & \textbf{100.00} $\vert$ \textbf{0.10}\,\,\,\,\,\, \\
& & GraphSAGE & 81.33 & \textbf{100.00} $\vert$ \textbf{0.08}\,\,\,\,\,\, & 84.34 $\vert$ 1.32\,\,\, & 96.39 $\vert$ 0.23\,\,\, & 98.80 $\vert$ 0.21\,\,\, \\
\cline{2-8}
& & GCN & 79.82 & 87.95 $\vert$ 1.22\,\,\, & 72.29 $\vert$ 2.23\,\,\, & 69.88 $\vert$ 1.38\,\,\, & \textbf{96.39} $\vert$ \textbf{0.84}\,\,\, \\
& RS & GAT & 78.92 & 84.34 $\vert$ 0.83\,\,\, & 71.08 $\vert$ 3.42\,\,\, & 66.27 $\vert$ 0.67\,\,\, & \textbf{95.18} $\vert$ \textbf{0.05}\,\,\, \\
& & GraphSAGE & 79.22 & 83.13 $\vert$ 2.24\,\,\, & 80.72 $\vert$ 2.27\,\,\, & 69.88 $\vert$ 1.57\,\,\, & \textbf{93.98} $\vert$ \textbf{0.00}\,\,\, \\
\hline
\end{tabular}
\end{table*}

We assume that the adversary has access to only 1\% of the complete dataset. During the test phase, we strategically choose to add triggers to 25\% of the samples in the test set.

\textbf{Attack efficacy}. When evaluating our experimental results, according to the attack performance of various $\alpha$ and $T_{G}$ mentioned later, we set $\alpha=\text{0.2}$ and $T_{G}=\text{0.5}$. The outcomes of our experiments are presented in Table~\ref{tab2}. The table clearly illustrates the performance of ARBC and other attacks, as our ASR reaches over 96.43\%, while the CAD remains below 0.64\%. Additionally, it is worth noting that the GTA-t attack fails to achieve significant attack effectiveness, highlighting the inadequacy of the subgraph trigger topology.

\begin{figure}[!htb]
\centering
\includegraphics[width=0.45\textwidth]{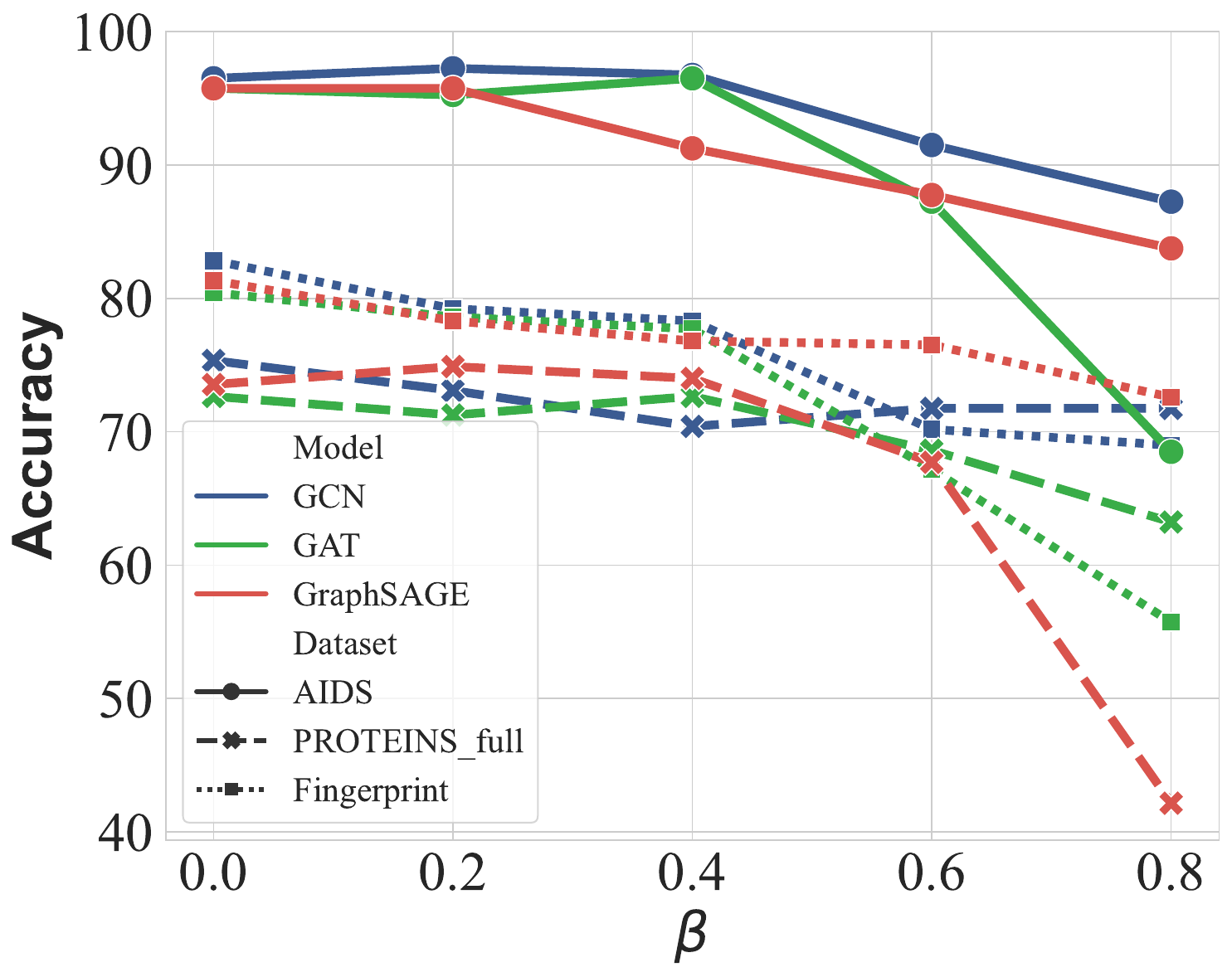}
\caption{The performance of Benign under various $\beta$ of RS on graph-level tasks.}
\label{Fig11}
\end{figure}

We applied the input-inspection backdoor mitigating strategy RS to defend against backdoor attacks. We define a parameter $\beta$, which means we randomly remove $\beta \times 100\%$ nodes of each graph sample, and for the rest nodes, we randomly set $\beta \times 100\%$ of their features to be 0. The larger $\beta$ is, the better RS can mitigate the impact of backdoor attacks. Figure~\ref{Fig11} shows the performance of Benign as $\beta$ varies from 0.0 to 0.8. We can observe that when $\beta=0.4$, the Benign model can maintain a good performance while mitigating backdoor attacks to the greatest extent. Therefore, we set $\beta=0.4$. Experimental results illustrate the robustness of our attack, which maintains a high ASR of approximately 96\% with only a marginal 2\% drop, while other attacks experience a significant 10\% to 30\% drop in their ASR.
	
\begin{figure*}[!htb]
\centering
\subfigure[AIDS]{
    \includegraphics[width=0.302\textwidth]{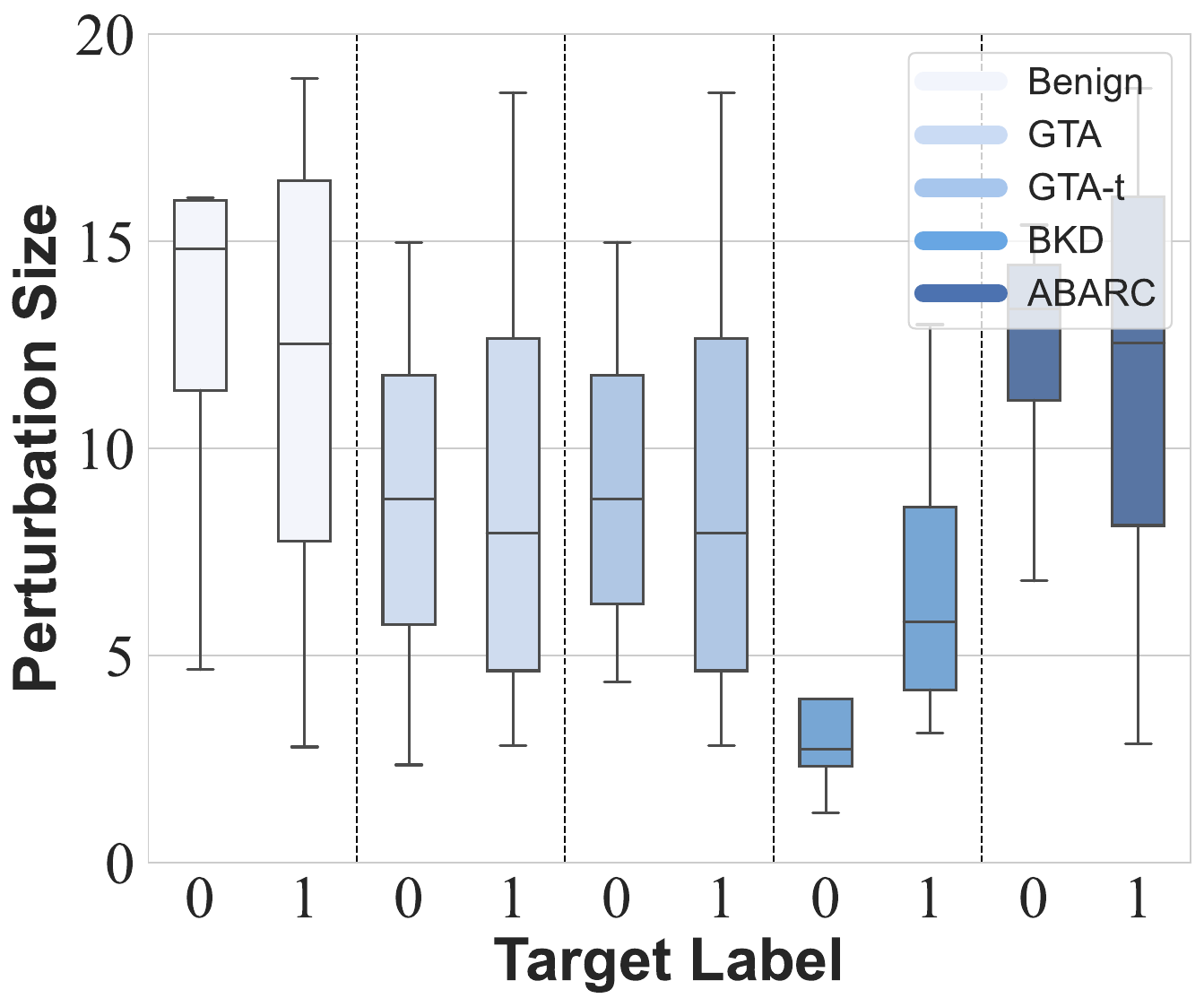}
}
\subfigure[PROTEINS\_full]{
    \includegraphics[width=0.31\textwidth]{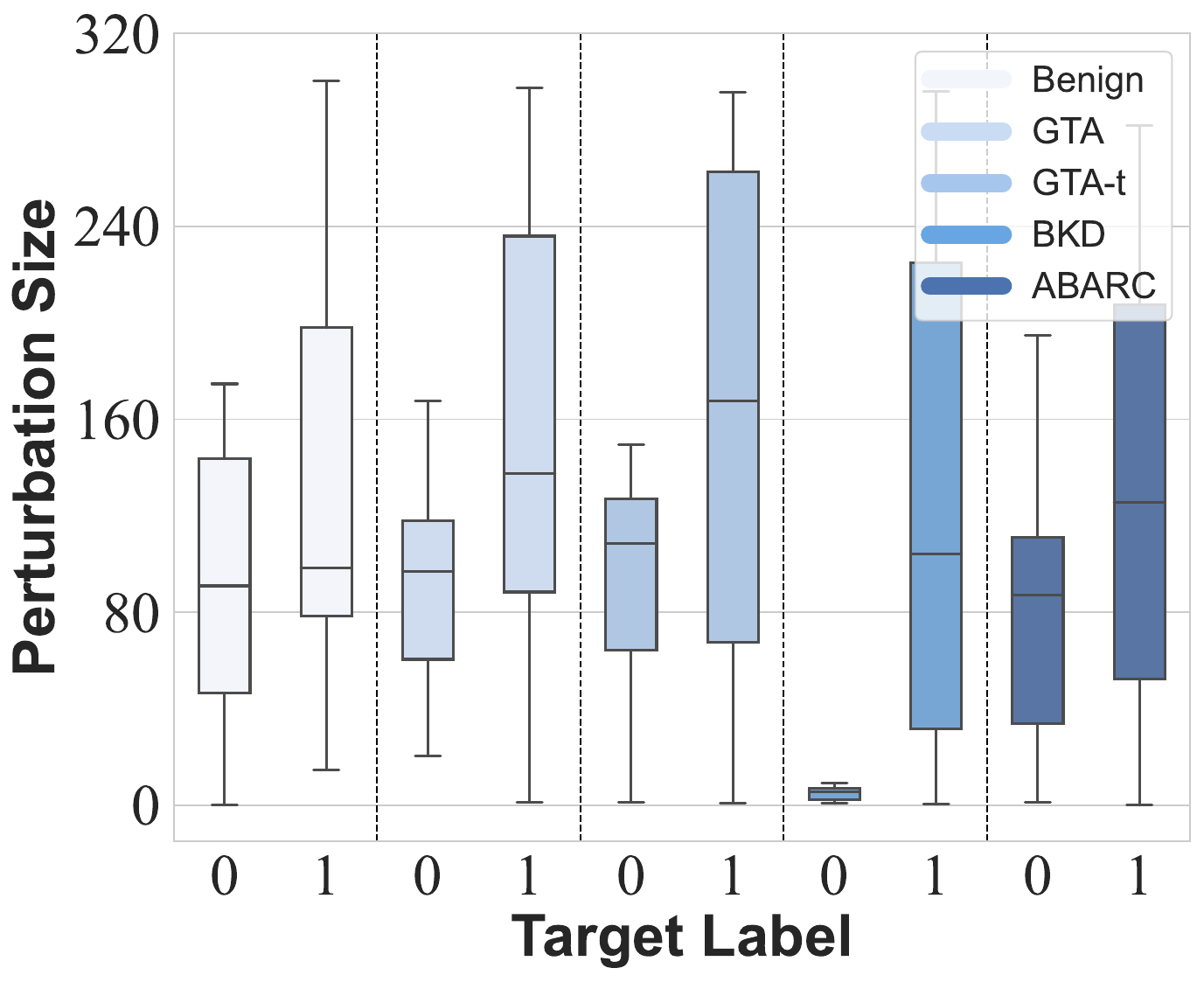}
}
    \subfigure[Fingerprint]{
    \includegraphics[width=0.302\textwidth]{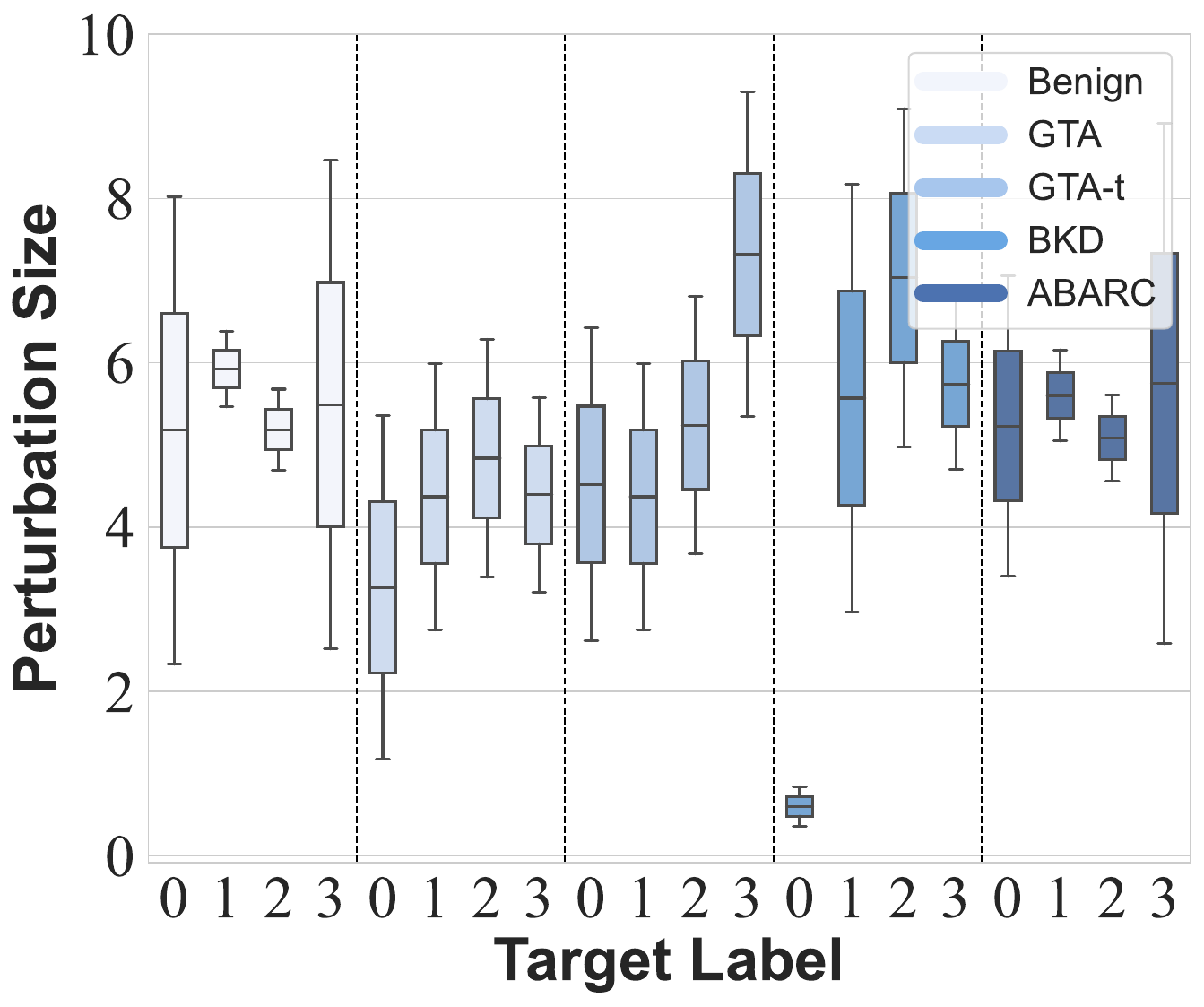}
}
\caption{Detection results of ABARC and other attacks by NC on the graph-level task using GCN.}
\label{Fig6}
\end{figure*}

We also apply the model-inspection backdoor defense strategy NC to evaluate the robustness of our graph-level backdoor attack. When performing NC detection, we assume that a specific class is the target class of the adversary and then add the same perturbation to the features of each node of all non-target class graph samples, looking for a model to classify all non-target class graph samples as a minimal perturbation of the target class. We generate a minimum perturbation for each class and determine whether the model contains a backdoor by comparing whether there are outliers in the generated perturbations. The results are shown in Figure~\ref{Fig6}. We can observe that BKD shows significant differences across the two classes. At the same time, the results of Benign and ours seem the most similar, implying the robustness of our graph-level backdoor attack.

\begin{figure*}[!htb]
\centering
\subfigure[AIDS]{
    \includegraphics[width=0.31\textwidth]{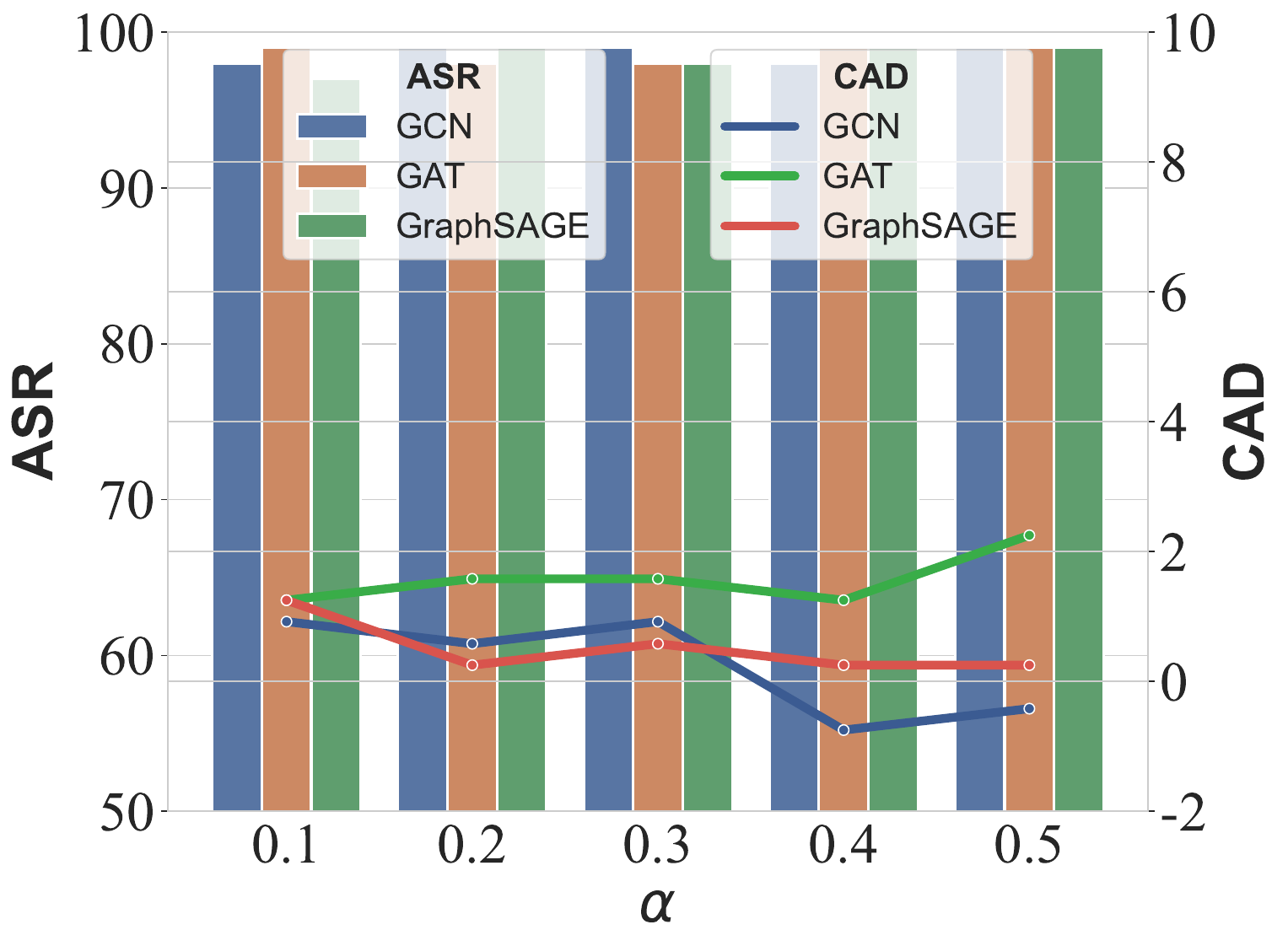}
}
\subfigure[PROTEINS\_full]{
    \includegraphics[width=0.31\textwidth]{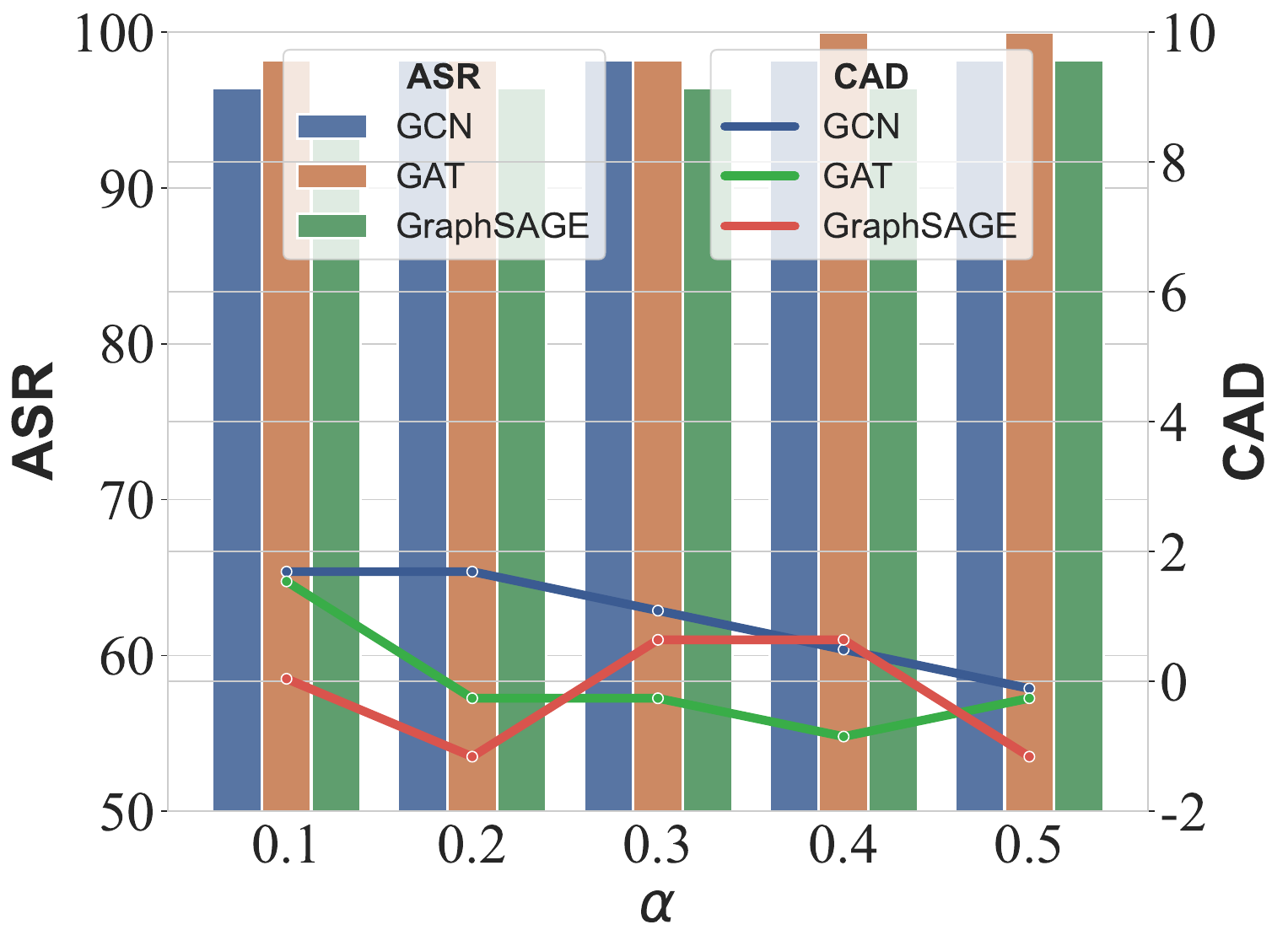}
}
    \subfigure[Fingerprint]{
    \includegraphics[width=0.31\textwidth]{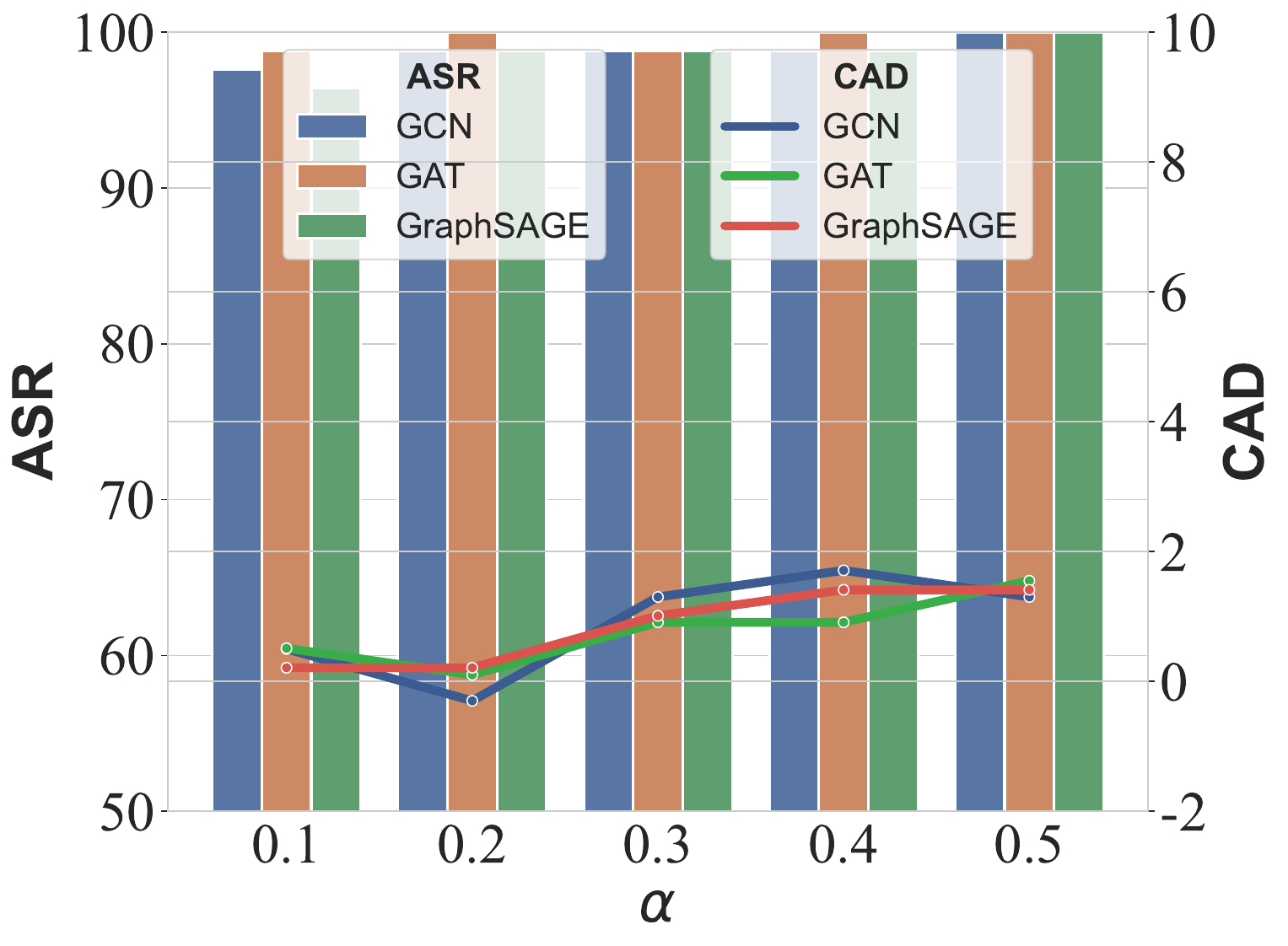}
}
\caption{The attack performance of ABARC under various trigger node proportion $\alpha$ on graph-level tasks when $T_{G}=\text{0.5}$.}
\label{Fig4}
\end{figure*}

\textbf{Trigger node proportion $\alpha$}. We conducted experiments by setting $T_{G}=\text{0.5}$. Figure~\ref{Fig4} depicts the performance of our method as $\alpha$ varies from 0.1 to 0.5. Notably, our attack displays robustness against changes in $\alpha$, with minimal impact on ASR and CAD. This resilience could be attributed to the fact that the dataset comprises graphs of varying sizes, and we ensure the adaptiveness of our designed triggers by selecting them in proportion to the graph size.

\begin{figure*}[!htb]
\centering
\subfigure[AIDS]{
    \includegraphics[width=0.31\textwidth]{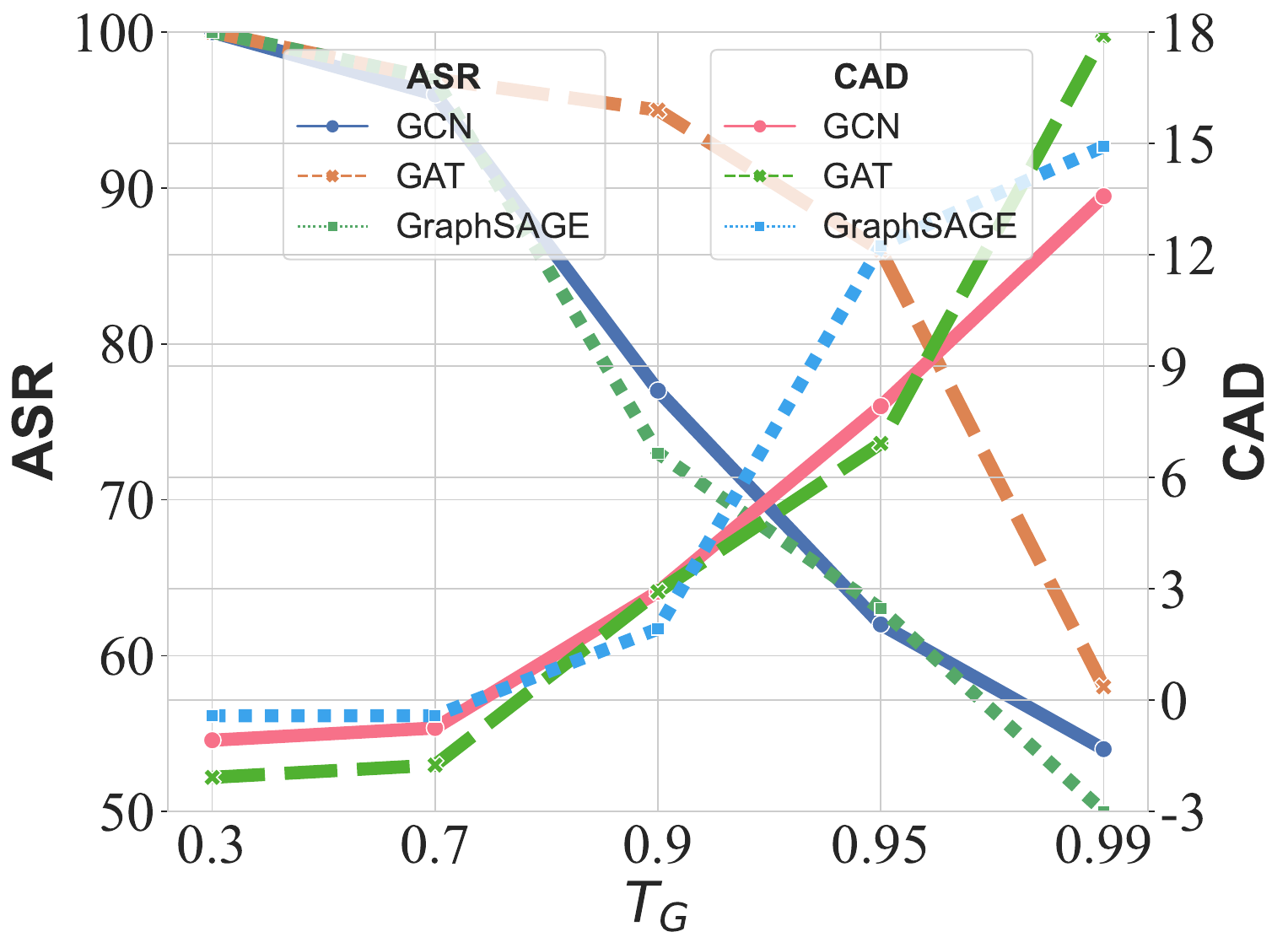}
}
\subfigure[PROTEINS\_full]{
    \includegraphics[width=0.31\textwidth]{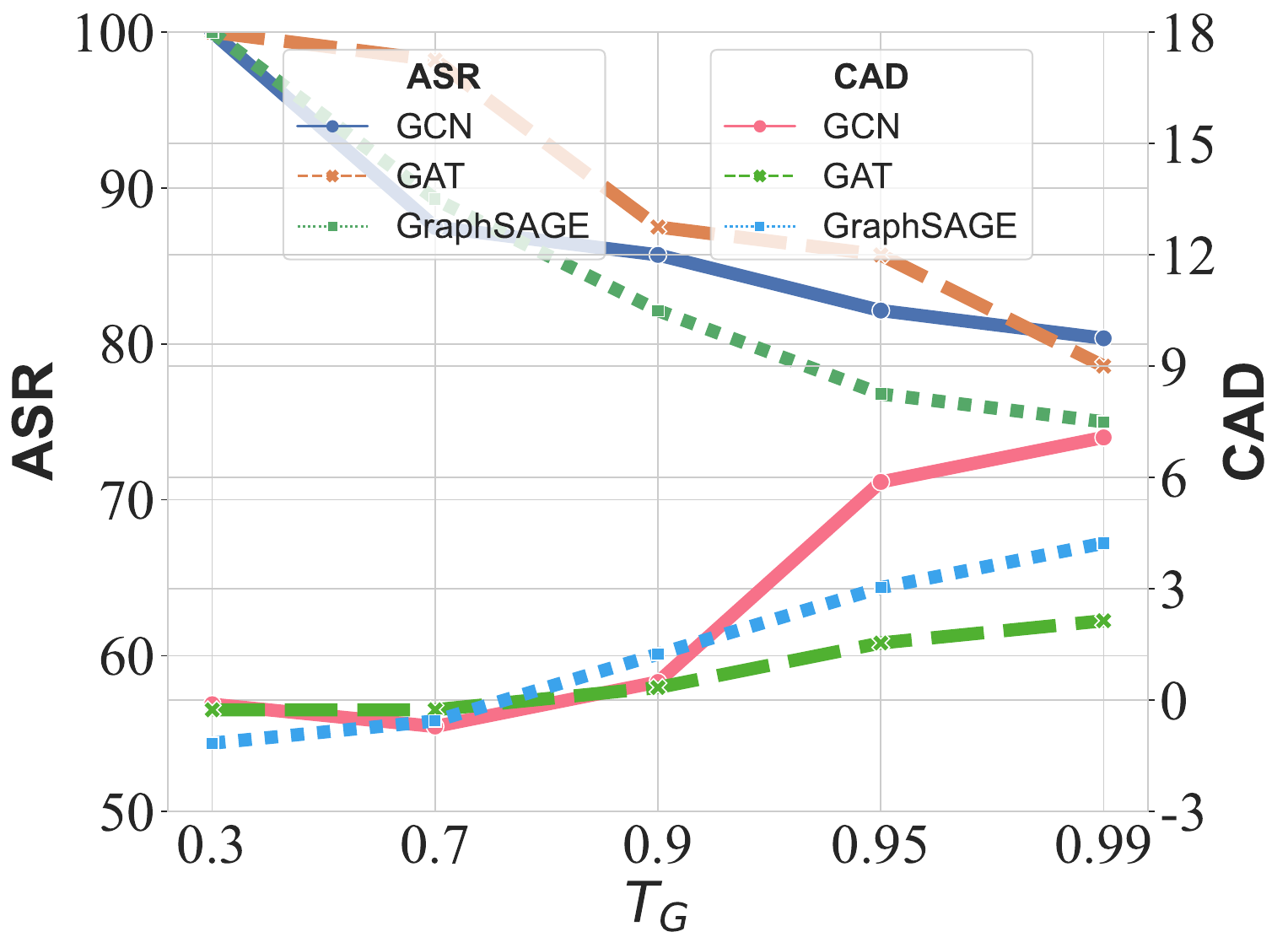}
}
    \subfigure[Fingerprint]{
    \includegraphics[width=0.31\textwidth]{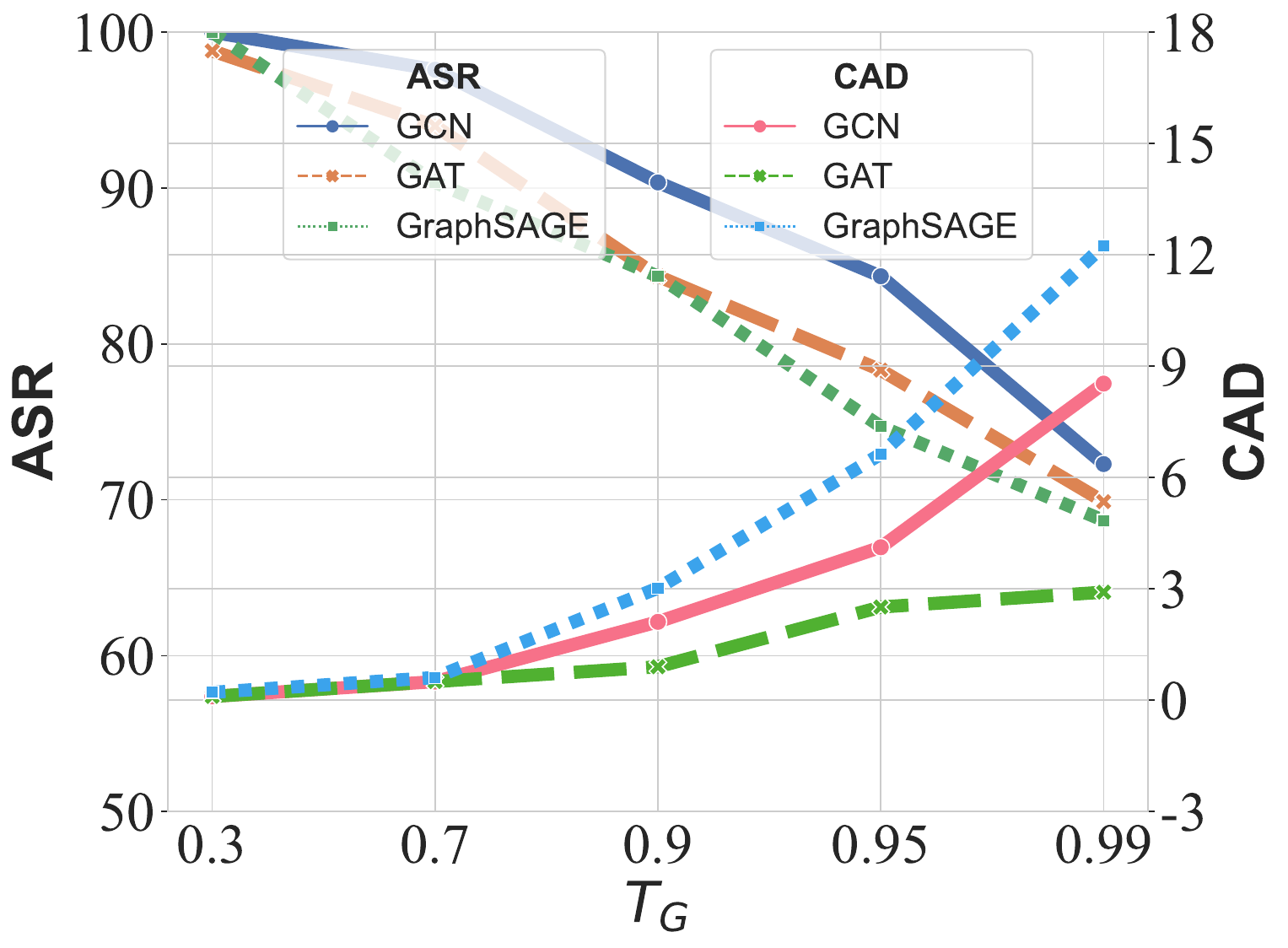}
}
\caption{The attack performance of ABARC under various similarity $T_{G}$ on graph-level tasks when $\alpha=\text{0.2}$.}
\label{Fig5}
\end{figure*}

\textbf{Similarity threshold $T_{G}$}. Furthermore, we investigate the impact of the similarity threshold $T_{G}$ by setting $\alpha=\text{0.2}$. Figure~\ref{Fig5} shows the performance of our method as the similarity threshold $T_{G}$ ranges from 0.3 to 0.99. The experimental results indicate that as the similarity between the original graph and the trigger-embedded graph increases, the ASR exhibits a noticeable decreasing trend, while CAD demonstrates a distinct increasing trend. Essentially, a higher similarity implies that the trigger becomes more evident as we shift the features of the trigger node, resulting in a more pronounced attack effectiveness.

\subsection{Results for Node Classification}

\renewcommand{\arraystretch}{1.4}
\begin{table*}[t]
\centering
\tabcolsep=0.45cm
\caption{Backdoor attack results of node-level tasks (ASR (\%) $\vert$ CAD (\%)).}
\label{tab3}
\begin{tabular}{cccccccc}
\hline
Dataset &Defense & Model & Benign & EXP & GTA & UGBA & ABARC (Ours)\\
\hline
\hline
\multirow{6}{*}{Cora} & & GCN & 85.79 & 73.80 $\vert$ 20.11 & 95.94 $\vert$ 1.66\,\,\, & 93.73 $\vert$ 1.29\,\,\, & \textbf{95.94} $\vert$ \textbf{0.74}\,\,\, \\
& None & GAT & 86.16 & 74.54 $\vert$ 20.48 & 96.31 $\vert$ 3.13\,\,\, & 94.83 $\vert$ 2.77\,\,\, & \textbf{98.89} $\vert$ \textbf{1.11}\,\,\, \\
& & GraphSAGE & 85.61 & 83.76 $\vert$ 6.27\,\,\, & \textbf{99.26} $\vert$ \textbf{0.74}\,\,\, & 98.52 $\vert$ 0.37\,\,\, & 98.16 $\vert$ 1.85\,\,\, \\
\cline{2-8}
& & GCN & 85.42 & 62.36 $\vert$ 1.66\,\,\, & 83.39 $\vert$ 2.03\,\,\, & 83.03 $\vert$ 1.29\,\,\, & \textbf{100.00} $\vert$ \textbf{0.55}\,\,\,\,\,\, \\
& RS & GAT & 84.31 & 63.10 $\vert$ 2.39\,\,\, & 80.44 $\vert$ 0.00\,\,\, & 73.80 $\vert$ 0.00\,\,\, & \textbf{99.26} $\vert$ \textbf{0.00}\,\,\, \\
& & GraphSAGE & 86.34 & 88.19 $\vert$ 1.47\,\,\, & 86.72 $\vert$ 2.21\,\,\, & 85.24 $\vert$ 1.10\,\,\, & \textbf{99.63} $\vert$ \textbf{0.00}\,\,\, \\
\hline
\multirow{6}{*}{CiteSeer} & & GCN & 76.73 & 40.54 $\vert$ 7.36\,\,\, & 96.40 $\vert$ 3.16\,\,\, & 96.70 $\vert$ 4.66\,\,\, & \textbf{99.40} $\vert$ \textbf{1.65}\,\,\, \\
& None & GAT & 77.33 & 40.24 $\vert$ 6.16\,\,\, & 94.29 $\vert$ 3.46\,\,\, & 98.50 $\vert$ 3.16\,\,\, & \textbf{100.00} $\vert$ \textbf{0.75}\,\,\,\,\,\, \\
& & GraphSAGE & 76.58 & 51.95 $\vert$ 1.50\,\,\, & \textbf{99.70} $\vert$ \textbf{0.30}\,\,\, & 94.89 $\vert$ 0.60\,\,\, & 99.70 $\vert$ 1.81\,\,\, \\
\cline{2-8}
& & GCN & 75.98 & 35.14 $\vert$ 1.21\,\,\, & 86.79 $\vert$ 2.71\,\,\, & 80.48 $\vert$ 2.71\,\,\, & \textbf{100.00} $\vert$ \textbf{0.00}\,\,\,\,\,\, \\
& RS & GAT & 75.68 & 31.53 $\vert$ 0.00\,\,\, & 80.78 $\vert$ 3.01\,\,\, & 72.67 $\vert$ 1.81\,\,\, & \textbf{100.00} $\vert$ \textbf{0.00}\,\,\,\,\,\, \\
& & GraphSAGE & 76.58 & 55.56 $\vert$ 0.60\,\,\, & 89.19 $\vert$ 0.00\,\,\, & 81.98 $\vert$ 2.11\,\,\, & \textbf{100.00} $\vert$ \textbf{0.30}\,\,\,\,\,\, \\
\hline
\multirow{6}{*}{Flickr} & & GCN & 49.86 & 50.42 $\vert$ 3.46\,\,\, & 95.24 $\vert$ 2.23\,\,\, & \textbf{97.06} $\vert$ \textbf{0.82}\,\,\, & 95.85 $\vert$ 1.09\,\,\, \\
& None & GAT & 49.77 & 44.82 $\vert$ 3.18\,\,\, & 96.72 $\vert$ 2.64\,\,\, & 97.84 $\vert$ 4.23\,\,\, & \textbf{99.72} $\vert$ \textbf{0.34}\,\,\, \\
& & GraphSAGE & 49.83 & 49.81 $\vert$ 0.75\,\,\, & \textbf{99.98} $\vert$ \textbf{0.23}\,\,\, & 99.94 $\vert$ 0.12\,\,\, & 99.88 $\vert$ 0.56\,\,\, \\
\cline{2-8}
& & GCN & 48.18 & 46.06 $\vert$ 1.35\,\,\, & 86.53 $\vert$ 3.25\,\,\, & 85.28 $\vert$ 2.33\,\,\, & \textbf{100.00} $\vert$ \textbf{0.00}\,\,\,\,\,\, \\
& RS & GAT & 47.62 & 39.45 $\vert$ 0.67\,\,\, & 84.27 $\vert$ 1.13\,\,\, & 73.06 $\vert$ 0.87\,\,\, & \textbf{99.87} $\vert$ \textbf{0.02}\,\,\, \\
& & GraphSAGE & 47.90 & 39.72 $\vert$ 1.22\,\,\, & 87.39 $\vert$ 0.55\,\,\, & 81.79 $\vert$ 1.85\,\,\, & \textbf{100.00} $\vert$ \textbf{0.00}\,\,\,\,\,\, \\
\hline
\end{tabular}
\end{table*}



We assume that the adversary has access to 5\% of the entire dataset. At test time, we add triggers to 50\% of the samples in the test set.

\textbf{Attack efficacy}. When evaluating our experimental results, according to the attack performance of various $\gamma$, $T_{N}$ and $T_{S}$ mentioned later, we set $\gamma=\text{0.3}$, $T_{N}=\text{0.5}$ and $T_{S}=\text{0.5}$. The outcomes of our experiments are presented in Table~\ref{tab3}. The table clearly illustrates the effectiveness of our method, as our ASR reaches over 95.94\%, while the CAD remains below 1.85\%.

\begin{figure}[!htb]
\centering
\includegraphics[width=0.45\textwidth]{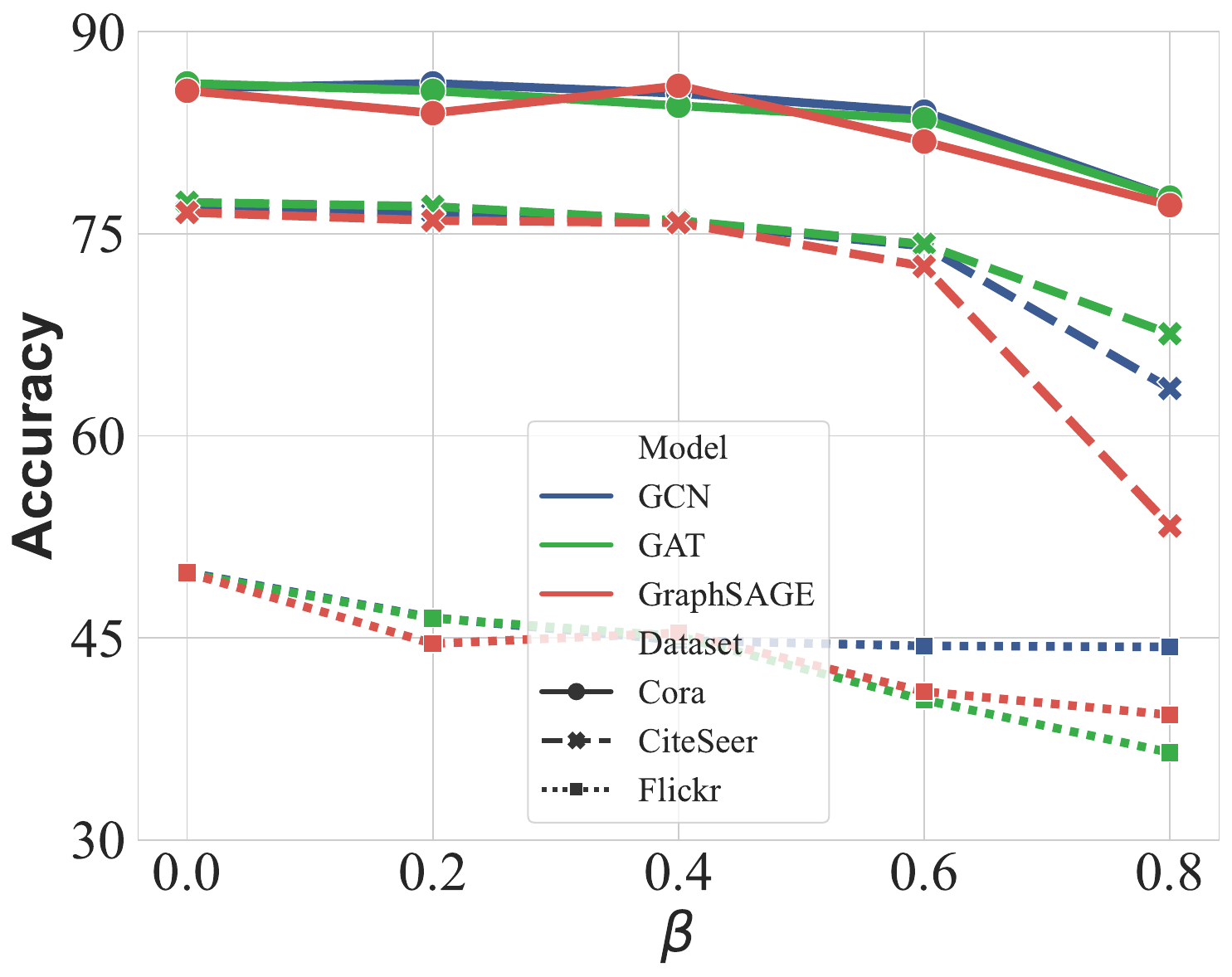}
\caption{The performance of Benign under various $\beta$ of RS on node-level tasks.}
\label{Fig12}
\end{figure}

We applied the input-inspection backdoor mitigating strategy RS to defend against backdoor attacks. Like our graph-level backdoor attack, we define a parameter $\beta$, which means we randomly prune the edge of the neighbor nodes of each node and set $\beta \times 100\%$ of each node's features to be 0. We test the performance of Benign as $\beta$ varies from 0.0 to 0.8, as shown in Figure~\ref{Fig12}. We can observe that when $\beta=0.4$, the Benign model can maintain a good performance while mitigating backdoor attacks to the greatest extent. Therefore, we set $\beta=0.4$. Our results illustrate the robustness of our method. Our attack maintains a high ASR of over 99\% with no drop, while other attacks experience a significant 10\% to 30\% drop in their ASR.

\begin{figure*}[!htb]
\centering
\subfigure[Cora]{
    \includegraphics[width=0.31\textwidth]{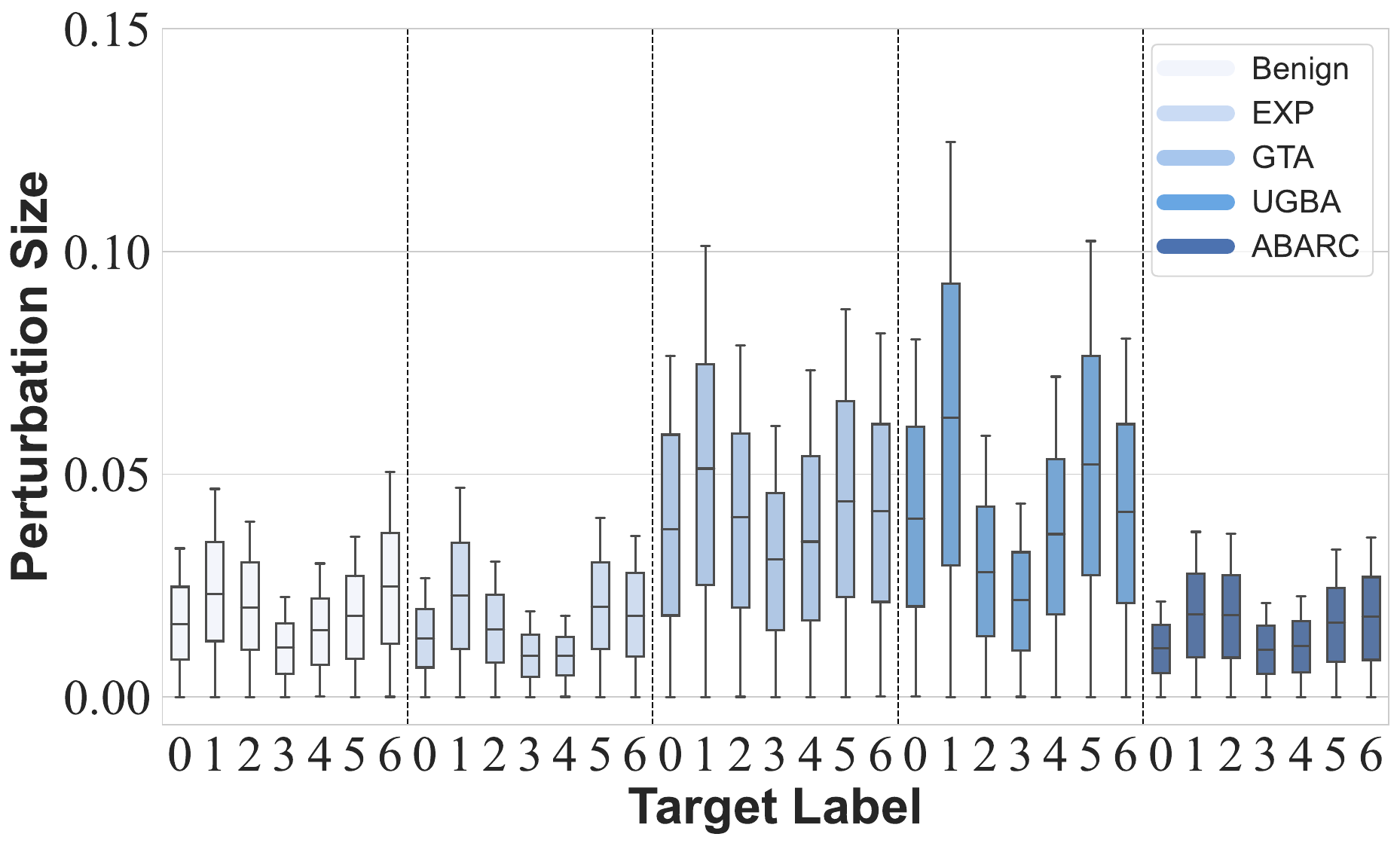}
}
\subfigure[CiteSeer]{
    \includegraphics[width=0.315\textwidth]{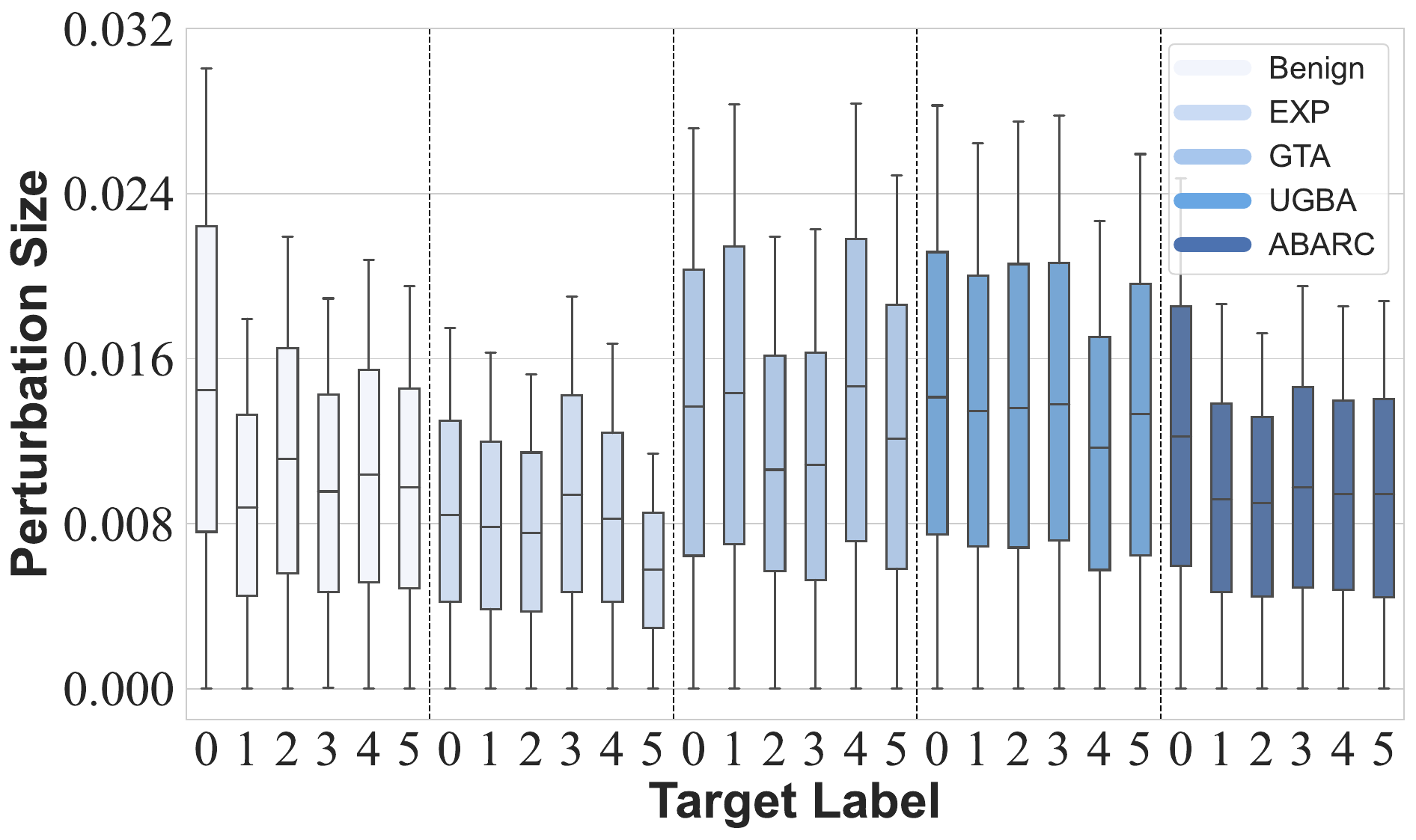}
}
    \subfigure[Flickr]{
    \includegraphics[width=0.31\textwidth]{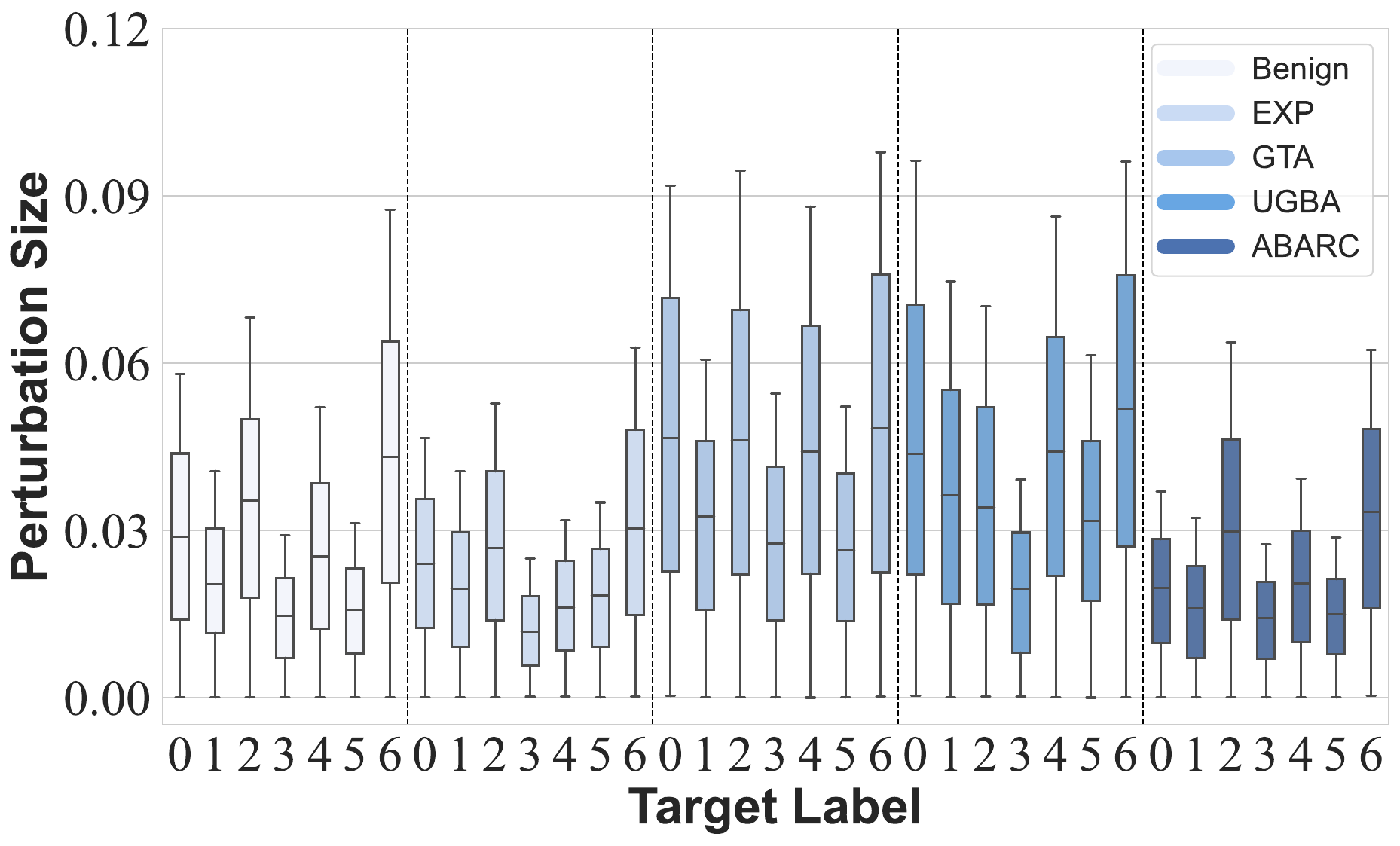}
}
\caption{Detection results of ABARC and other attacks by NC on the node-level task using GCN.}
\label{Fig9}
\end{figure*}

We also applied the model-inspection backdoor detection defense strategy NC to evaluate the robustness of our node-level backdoor attack. When performing NC detection, we assume that a specific class is the target class of the adversary and then add the same perturbation to the features of each non-target class node, looking for a model to classify all non-target class node samples as a minimal perturbation of the target class. We generate a minimum perturbation for each class and determine whether the model contains a backdoor by comparing whether there are outliers in the generated perturbations. The results are shown in Figure~\ref{Fig9}. We can observe that for all node-level attacks, NC cannot detect the backdoor. The detection results of our model are very similar to Benign, which reflects the robustness of our node-level backdoor attack.

\begin{figure*}[!htb]
\centering
\subfigure[Cora]{
    \includegraphics[width=0.31\textwidth]{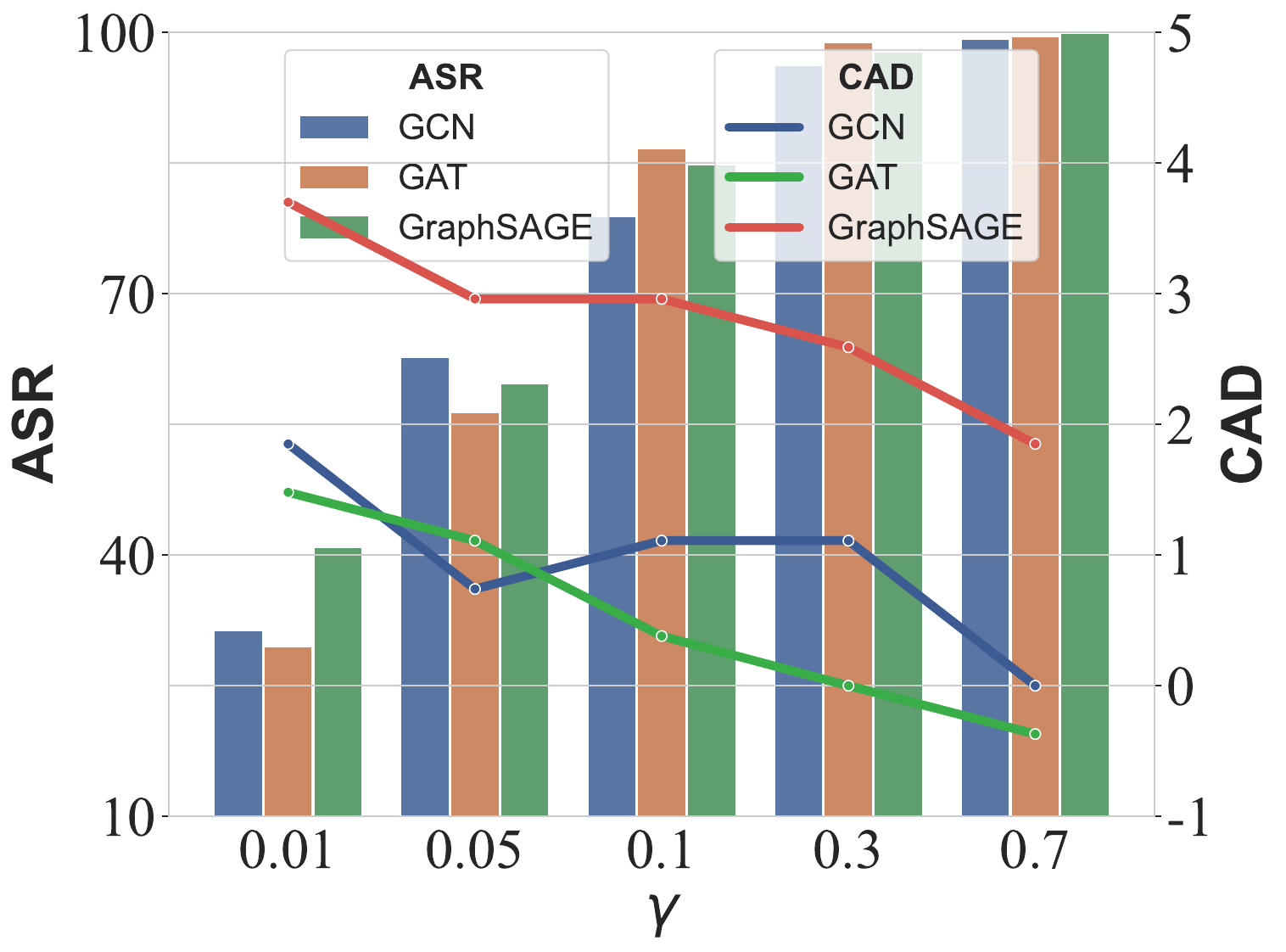}
}
\subfigure[CiteSeer]{
    \includegraphics[width=0.31\textwidth]{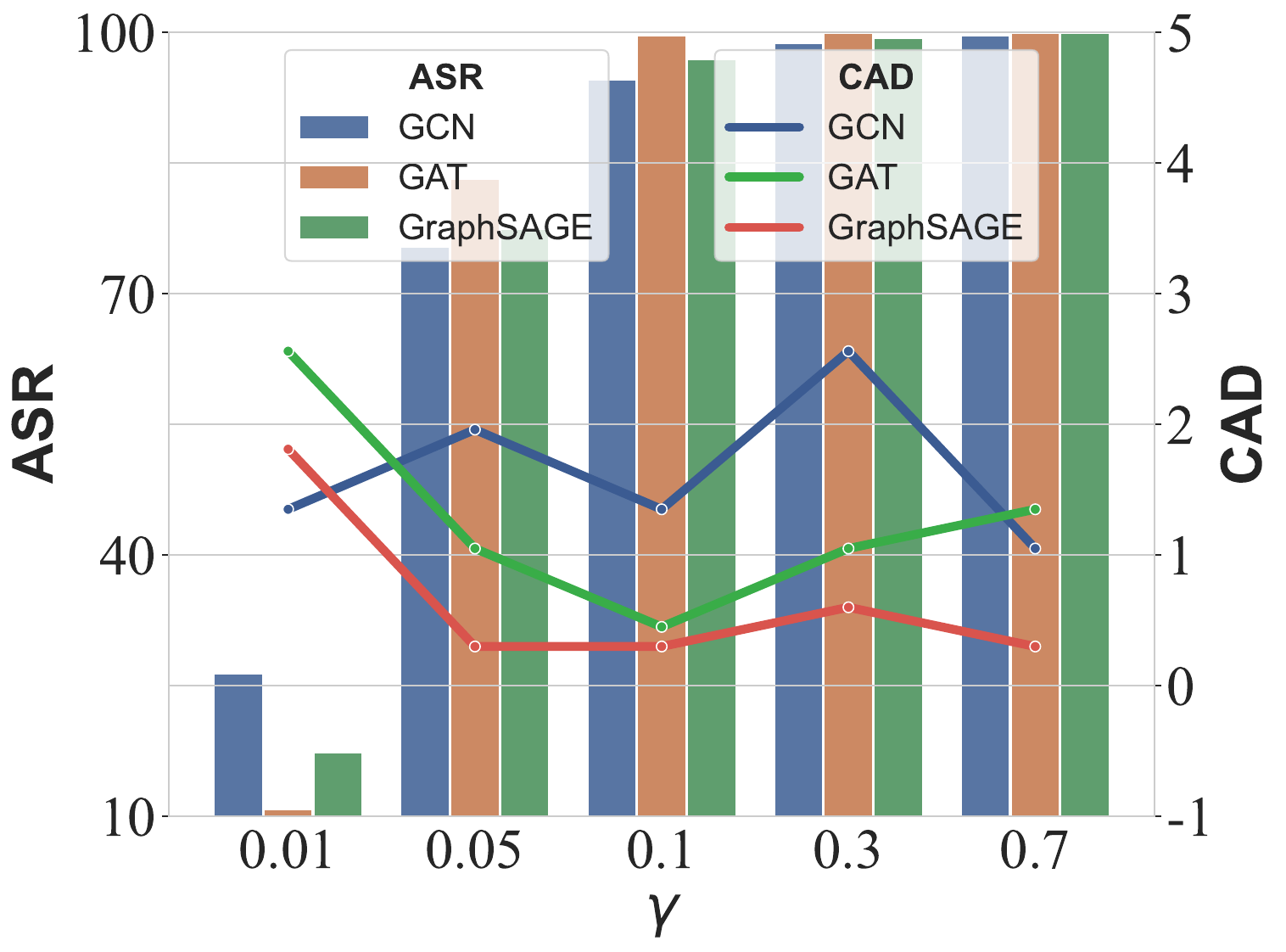}
}
    \subfigure[Flickr]{
    \includegraphics[width=0.31\textwidth]{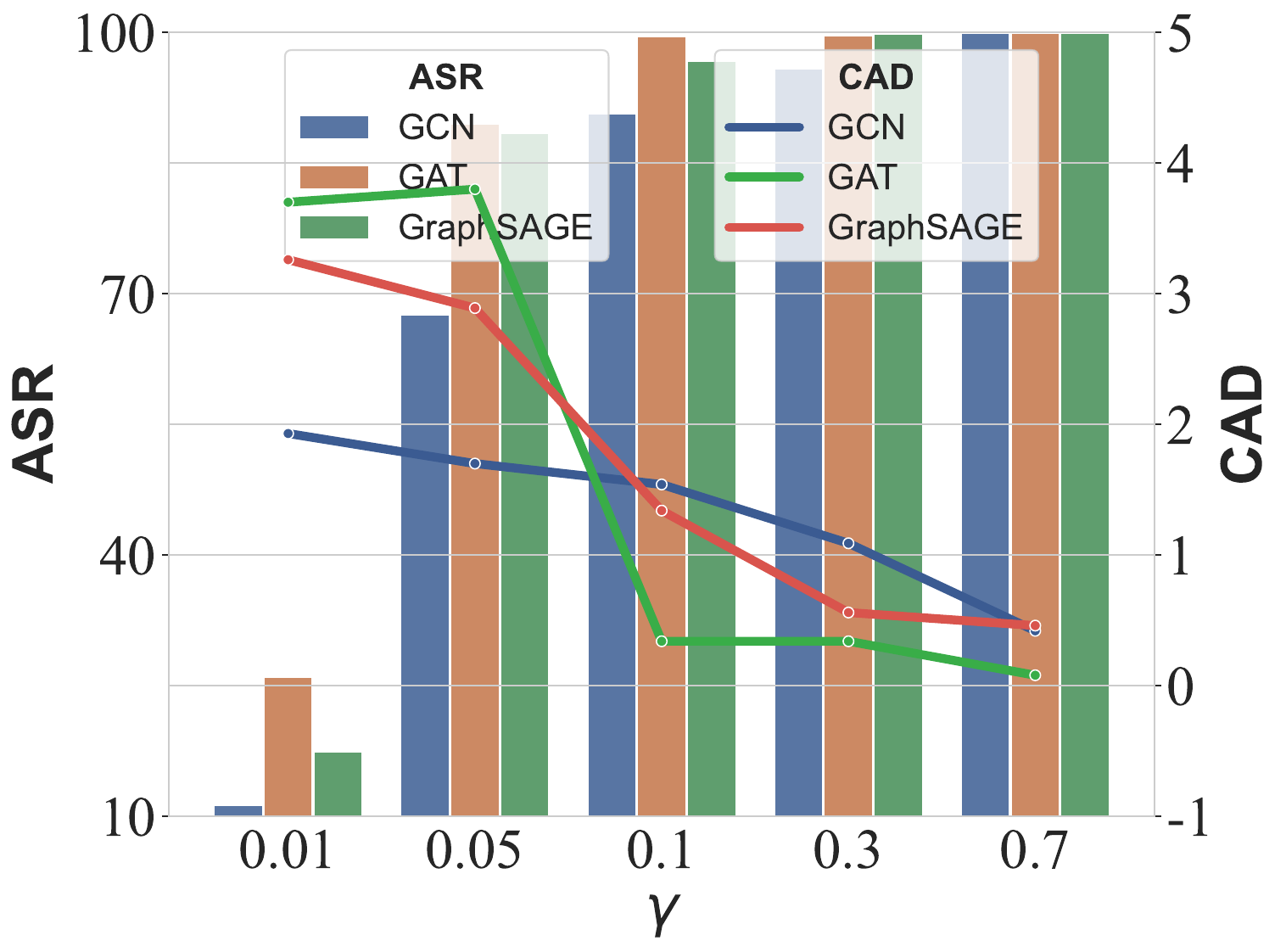}
}
\caption{The attack performance of ABARC under various feature trigger proportion $\gamma$ on node-level tasks when $T_{N}=\text{0.5}$ and $T_{S}=\text{0.5}$.}
\label{Fig7}
\end{figure*}

\textbf{Feature trigger proportion $\gamma$}. We set $T_{N}=\text{0.5}$ and $T_{S}=\text{0.5}$ to evaluate the impact of trigger node proportion $\gamma$ on our attack. Figure~\ref{Fig7} illustrates the performance of our method as $\gamma$ varies from 0.01 to 0.7. The experimental results indicate that as the proportion of feature triggers increases, the ASR exhibits a noticeable increasing trend, while CAD demonstrates a distinct decreasing trend.

\begin{figure*}[!htb]
\centering
\subfigure[Cora]{
    \includegraphics[width=0.31\textwidth]{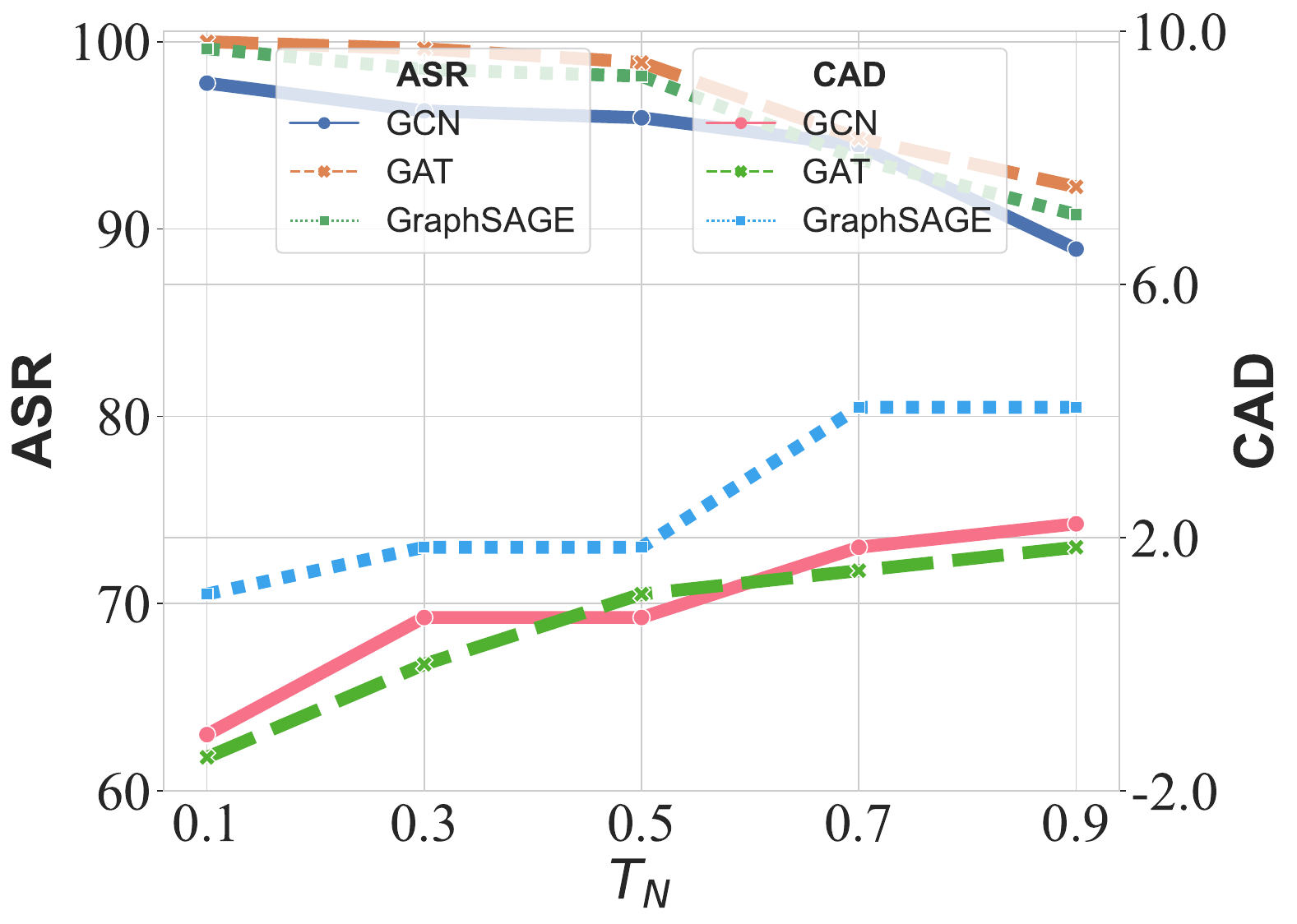}
}
\subfigure[CiteSeer]{
    \includegraphics[width=0.31\textwidth]{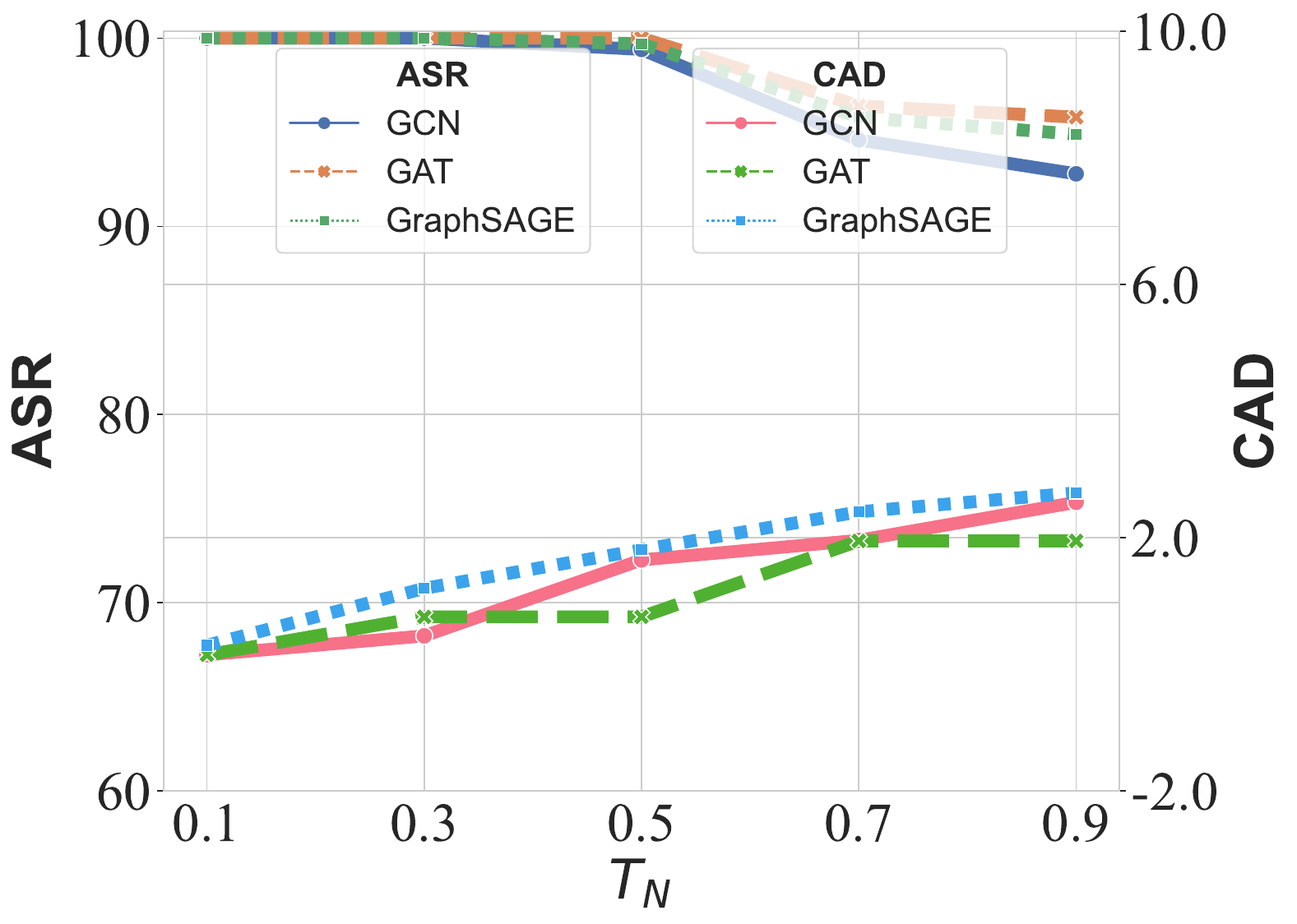}
}
    \subfigure[Flickr]{
    \includegraphics[width=0.31\textwidth]{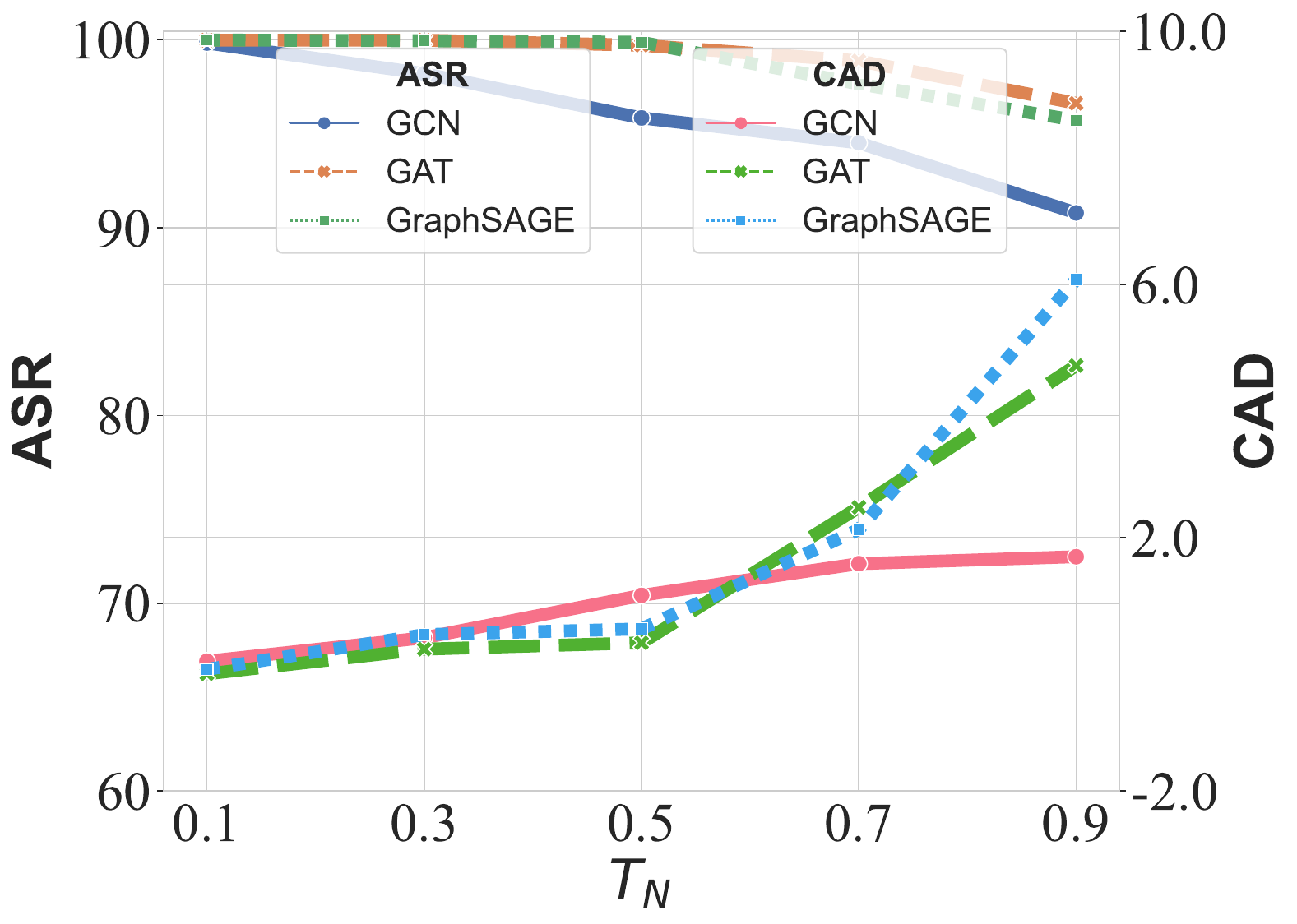}
}
\caption{The attack performance of ABARC under various similarity threshold $T_{N}$ on node-level tasks when $\gamma=\text{0.3}$ and $T_{S}=\text{0.5}$.}
\label{Fig8}
\end{figure*}

\textbf{Similarity threshold $T_{N}$}. We set $\gamma=\text{0.3}$ and $T_{S}=\text{0.5}$ to evaluate the impact of similarity threshold $T_{N}$ on our attack. Figure~\ref{Fig8} illustrates the performance of our method as the similarity threshold $T_{N}$ varies from 0.1 to 0.9. Notably, our attack method displays robustness against changes in $T_{N}$, with minimal impact on ASR and CAD. This phenomenon could be attributed to the dataset's high number of features per node. By selectively choosing a subset of features as triggers, any modifications made to them would have a minimal impact on the similarity between nodes before and after the changes.

\begin{figure*}[!htb]
\centering
\subfigure[Cora]{
    \includegraphics[width=0.31\textwidth]{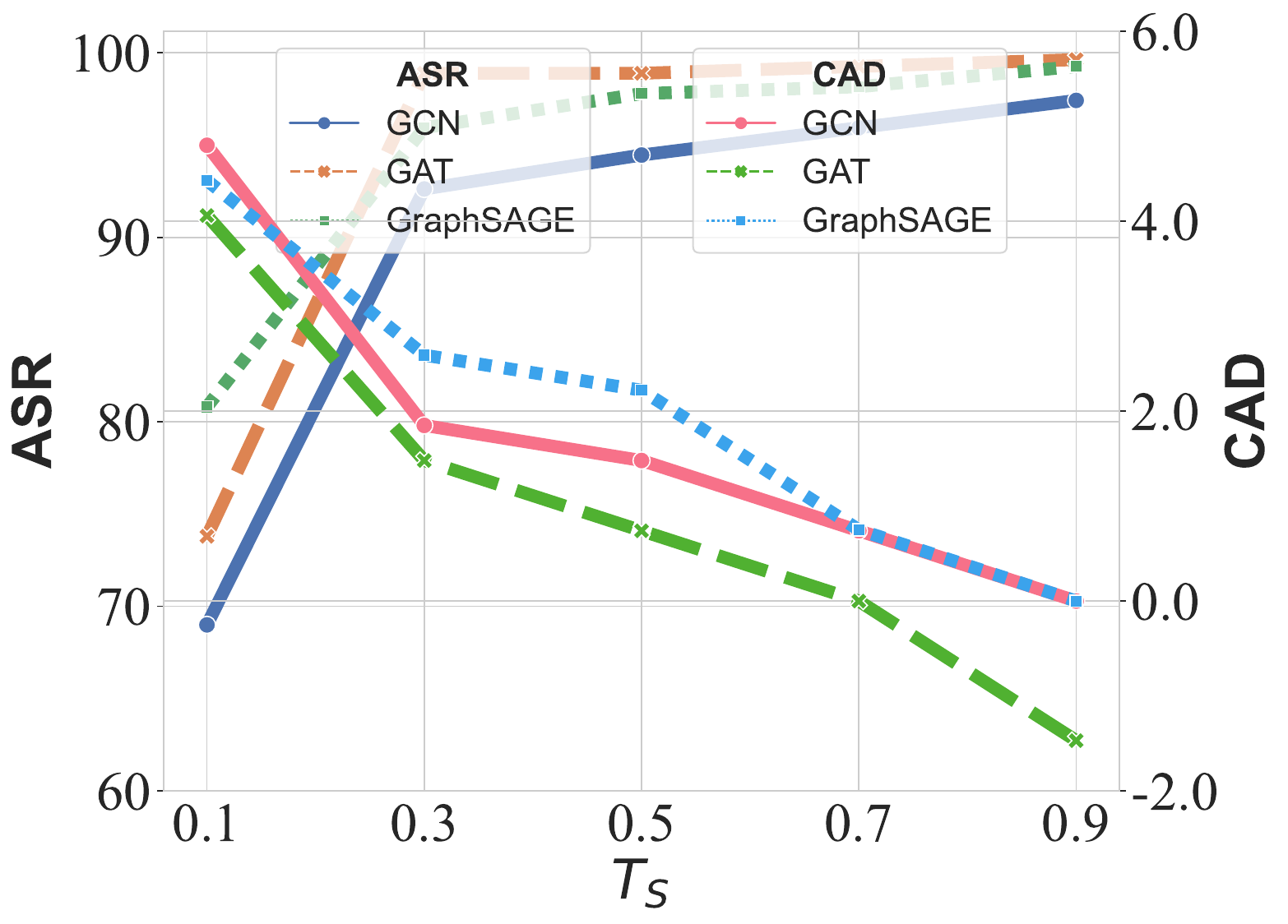}
}
\subfigure[CiteSeer]{
    \includegraphics[width=0.31\textwidth]{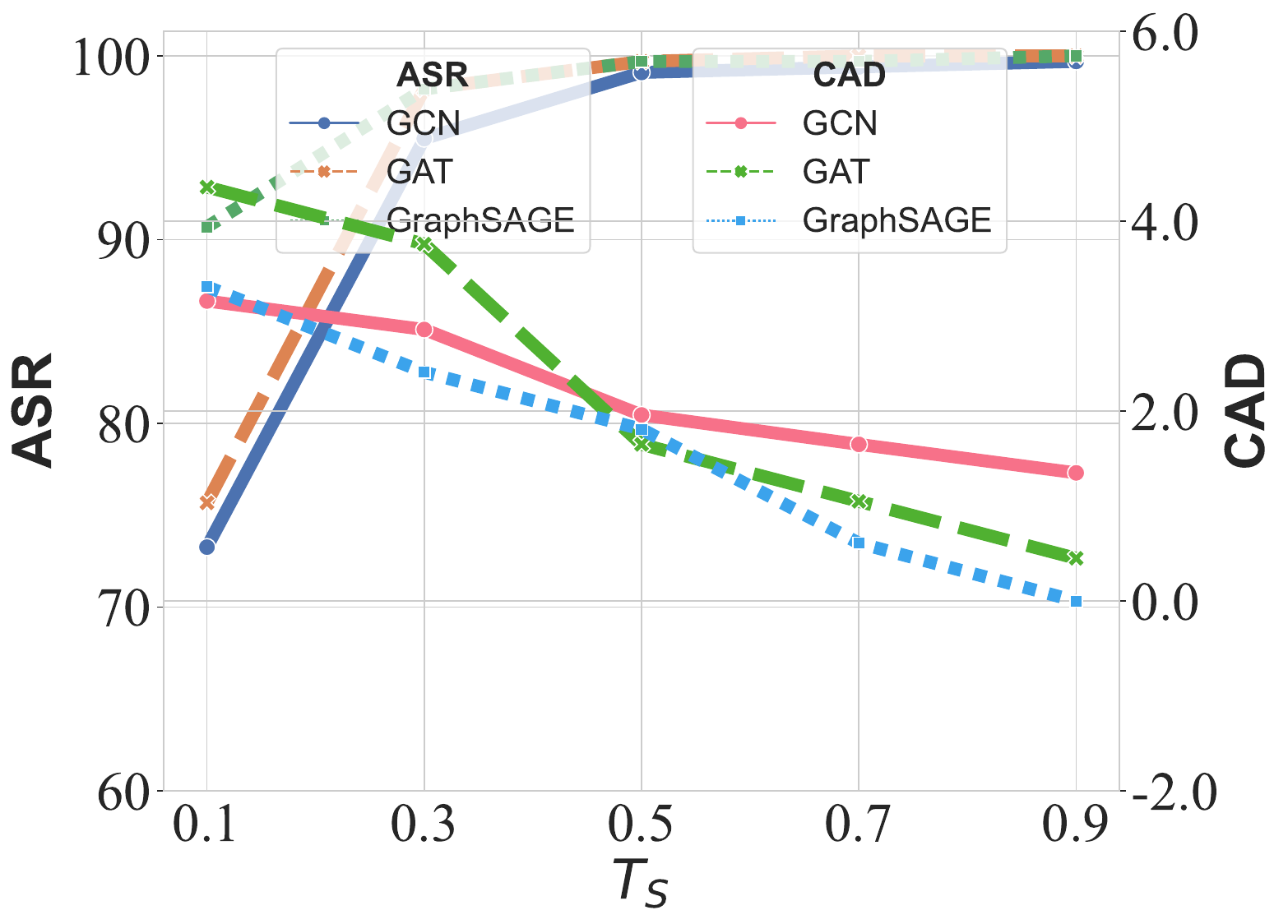}
}
    \subfigure[Flickr]{
    \includegraphics[width=0.31\textwidth]{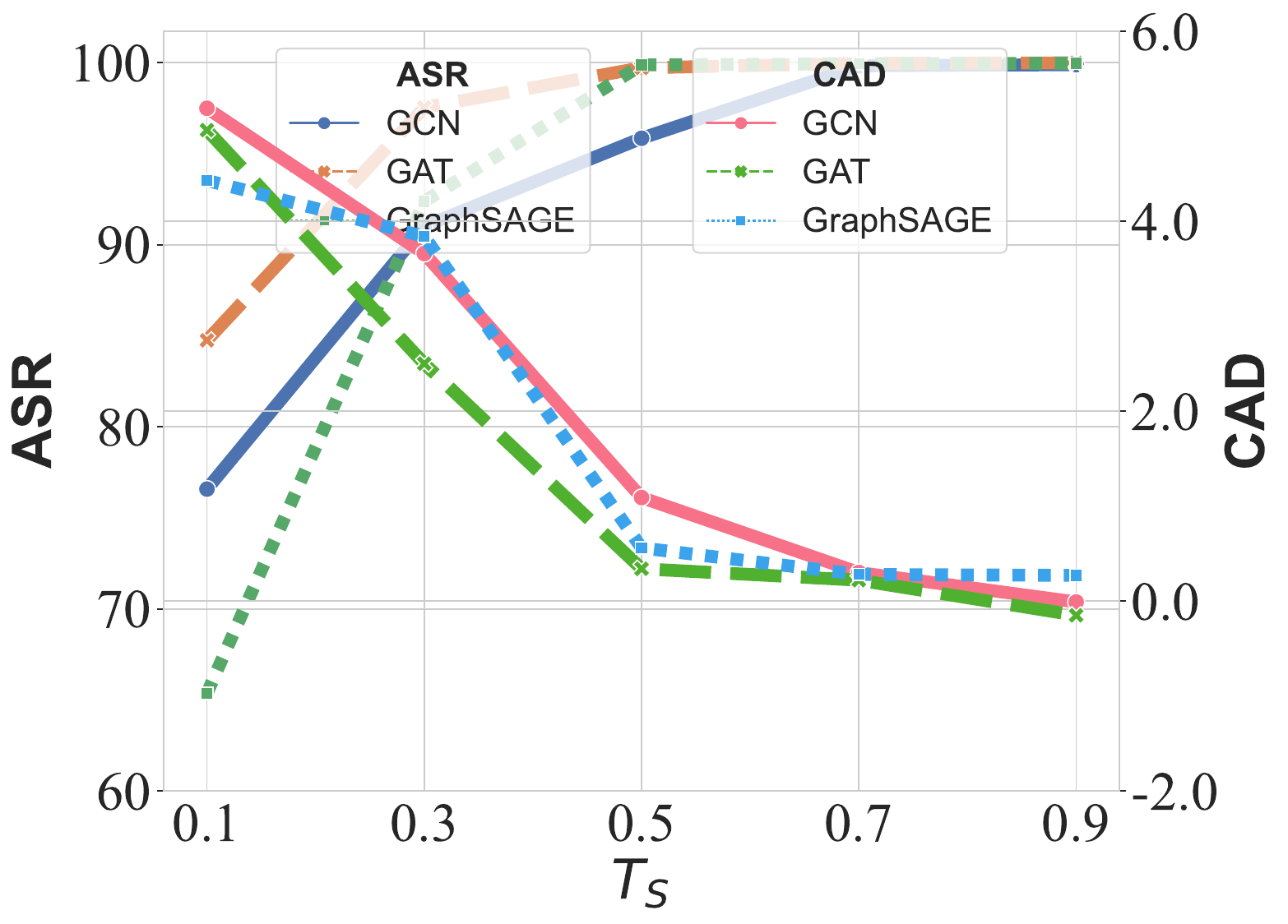}
}
    \caption{The attack performance of ABARC under various adaptive edge-pruning threshold $T_{S}$ on node-level tasks when $\gamma=\text{0.3}$ and $T_{G}=\text{0.5}$.}
\label{Fig10}
\end{figure*}

\textbf{Adaptive edge-pruning threshold $T_{S}$}. We set $\gamma=\text{0.3}$ and $T_{N}=\text{0.5}$ to evaluate the impact of edge-pruning threshold $T_{S}$ on our attack. Figure~\ref{Fig10} illustrates the performance of our node-level backdoor attack as $T_{S}$ varies from 0.1 to 0.9. The experimental results indicate that as the threshold increases, the ASR exhibits a noticeable increasing trend, while CAD demonstrates a distinct decreasing trend.

\subsection{The Performance of ABARC with Varying Graph Sizes}

The performance of ABARC has been thoroughly evaluated across a diverse set of graph samples varying significantly in size and complexity. The results demonstrate that the effectiveness of ABARC remains robust regardless of these variations.

In smaller graph samples, ABARC demonstrates high effectiveness due to the limited number of nodes and edges, which facilitates easier manipulation and more apparent impact of the adversarial perturbations.

As the graph size increases, the challenge of embedding the backdoor while maintaining evasiveness also increases. However, ABARC has been designed to scale effectively, ensuring that the attack remains potent even in larger graph samples. This is achieved by building a dynamic trigger pattern and adaptively selecting nodes or node features as triggers based on the graph sample node scale and node feature dimensions, maximizing the impact with minimal changes.

\section{Conclusion}

In conclusion, this paper addresses the limitations of fixed pattern triggers and the lack of reasonable constraints of backdoor attacks against GNNs for graph-level and node-level tasks. The proposed adaptive backdoor attacks, ABARC, effectively manipulate model behavior while maintaining evasiveness, highlighting the significance of security considerations in GNN applications.

\subsection{Limitations}

However, this paper still falls short in terms of a comprehensive semantic constraint design, resulting in a lack of evasiveness of the attacks. The constraints designed in this paper consider sample similarity, sample feature value range, and feature value type, ensuring the trigger's rationality to a certain extent. However, it is still a relatively broad constraint, lacking in-depth consideration of the meaning of graph sample nodes. For example, for molecular structure graph data, nodes represent atoms, and chemical bonds connect atoms. The valence electrons contained in the connected atomic locks should correspond. The dynamic trigger mode designed in this paper cannot achieve a high attack success rate under the premise of meeting this condition.

\subsection{Defense Methods}

ABARC can be defended from the following two aspects:

Defense can be carried out through \textbf{data filtering}. Since ABARC lacks comprehensive semantic constraints, defenders can filter graph data samples according to their characteristics before model training. However, this method relies on an in-depth understanding of graph data samples, and it is difficult to find a standard filtering method for various graph data. This complexity underscores the need for further research and the development of more effective filtering methods.

Furthermore, defense can also be carried out through \textbf{model comparison}. While ABARC ensures a high attack success rate, it minimizes the model's classification accuracy decline for normal data. However, the model's classification accuracy for normal data will still decline. Therefore, defenders can use their own reliable data sets to train a new model and determine whether the model has been attacked by comparing the classification accuracy of the new model and the suspicious model for normal data. Since the classification accuracy of the model attacked by ABARC for normal data has only slightly decreased, determining whether the suspicious model has been attacked remains a complex issue that requires further research. Simultaneously, better model comparison methods can also be further studied.

\subsection{Future Research}

There is a need for the development of more sophisticated and evasive dynamic trigger patterns, which could serve as a valuable avenue for future in-depth research. Furthermore, one of the future research directions could be the in-depth study of the working principle of GNNs to analyze and explain the security threats they face; at the same time, more effective and evasive backdoor attacks based on GNNs' inherent flaws could be developed. Another research direction could explore additional techniques to enhance the robustness of GNNs against backdoor attacks. Further investigation into advanced defense mechanisms and techniques to detect and mitigate backdoor attacks would be valuable. Moreover, exploring the impact of backdoor attacks on more complex graph structures, such as multi-relational graphs and heterogeneous graphs, would extend our understanding of their implications in real-world scenarios. Additionally, research into addressing backdoor attacks in federated learning settings, where GNN models are trained on distributed data sources, presents an interesting avenue for exploration. Overall, advancing research in GNN security will be crucial to safeguarding the integrity and reliability of graph-based applications in various domains. Besides, once ABARC is more thoroughly explored, its application or similar methodologies in other neural networks or against different types of attacks may be investigated.

\normalem
\bibliographystyle{IEEEtran}
\bibliography{main}

\begin{IEEEbiography}[{\includegraphics[width=1in,height=1.25in,clip,keepaspectratio]{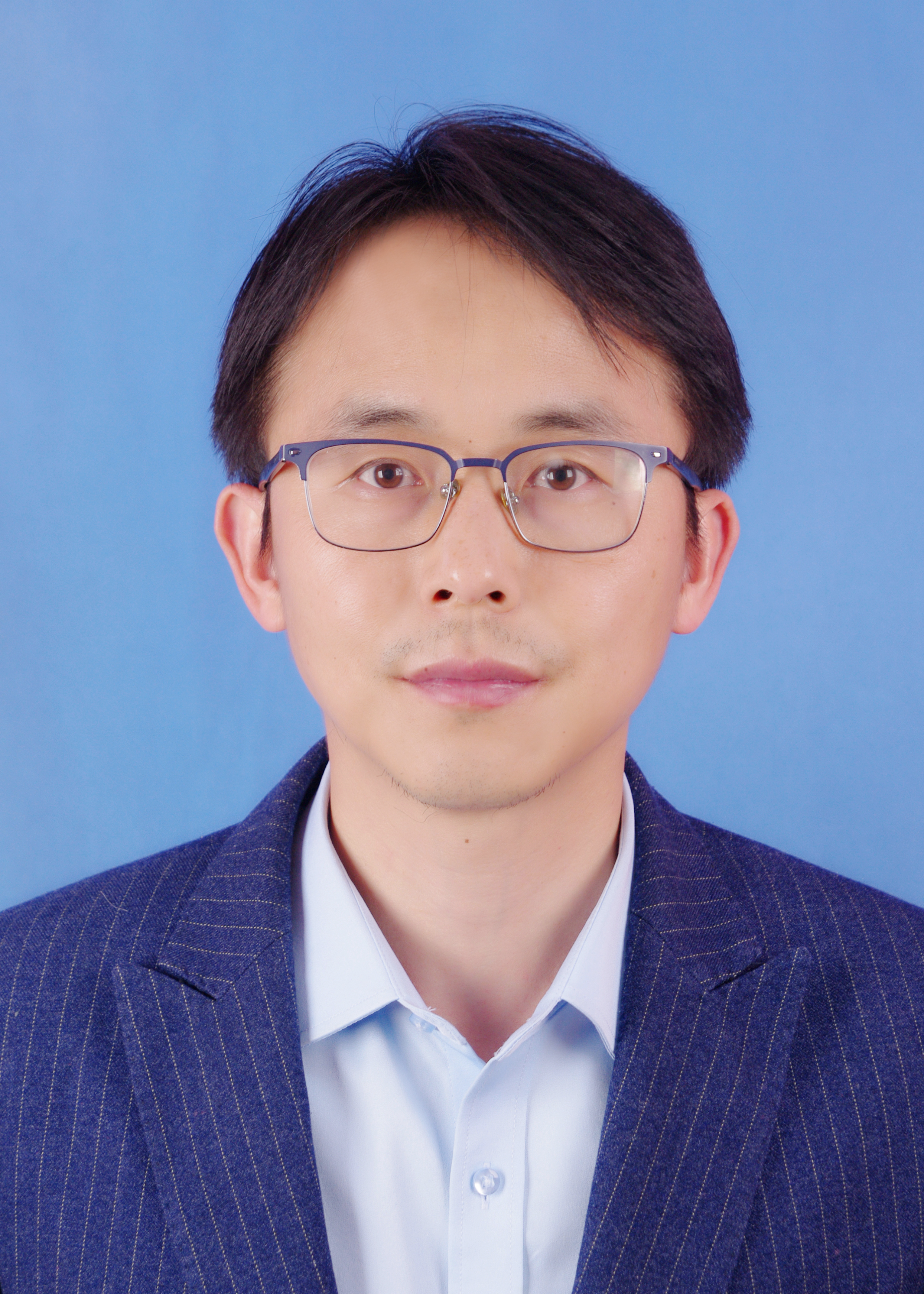}}]{Xuewen Dong} received his B.E., M.S.\ and Ph.D.\ degrees in Computer Science and Technology from the Xidian University, China, in 2003, 2006 and 2011, respectively. From 2016 to 2017, he was with the Oklahoma State University of USA as a visiting scholar. He is currently a professor in the School of Computer Science and Technology, Xidian University. His research interests include wireless network security and Blockchain.
\end{IEEEbiography}

\begin{IEEEbiography}[{\includegraphics[width=1in,height=1.25in,clip,keepaspectratio]{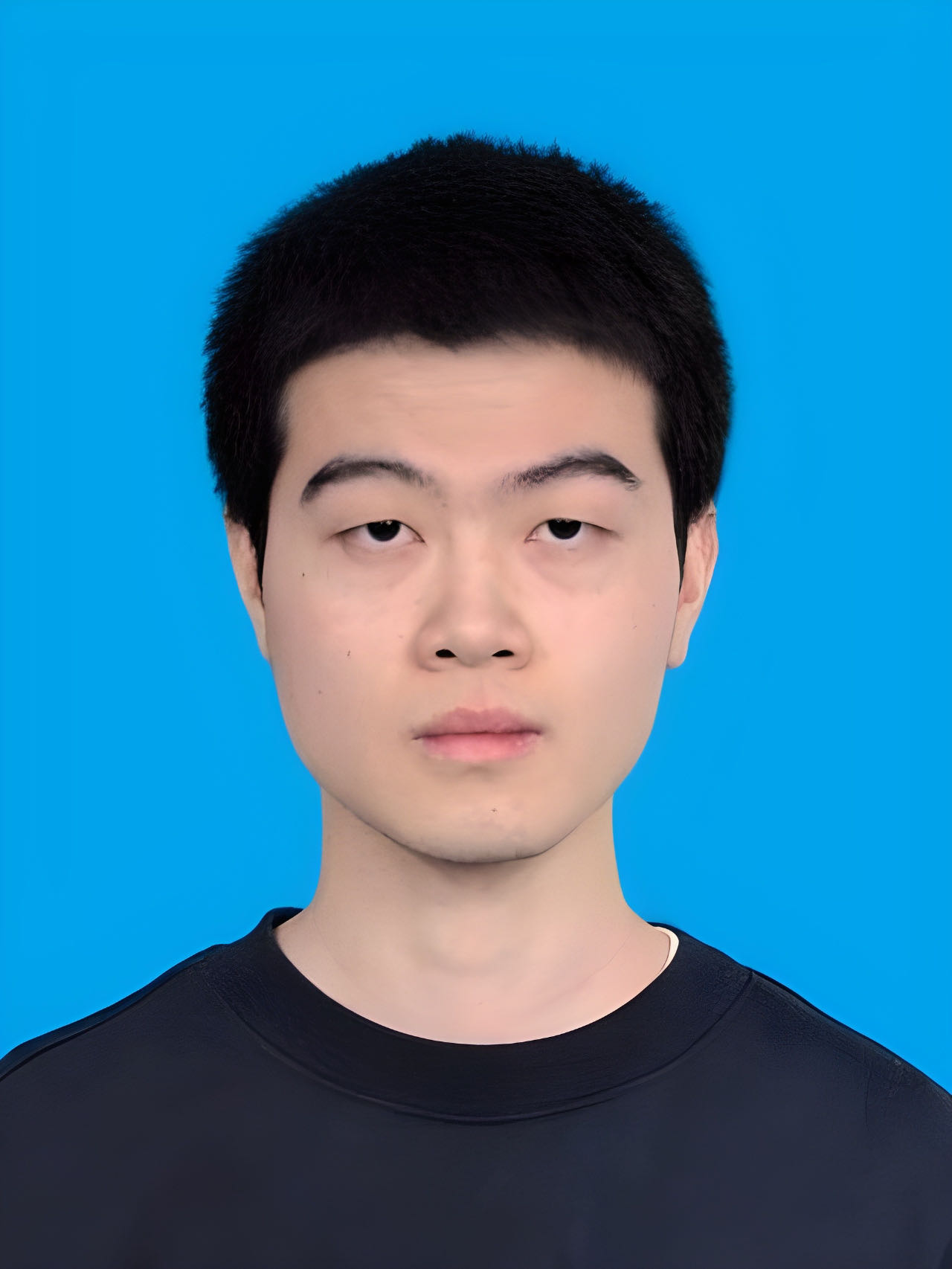}}]{Jiachen Li} received his B.E.\ and B.S.\ degrees in Computer Science and Technology from the Xidian University, China, in 2021 and 2024, respectively. His research interests include graph neural networks and backdoor attacks.
\end{IEEEbiography}

\begin{IEEEbiography}[{\includegraphics[width=1in,height=1.25in,clip,keepaspectratio]{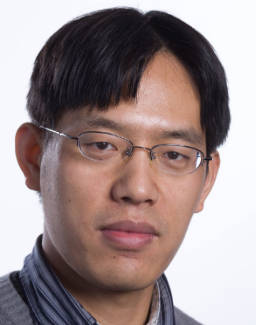}}]{Shujun Li} (M'2008, SM'2012) received his B.E.\ degree in Information Science and Engineering and his Ph.D.\ degree in Information and Communication Engineering from Xi'an Jiaotong University, China, in 1997 and 2003, respectively. He is Professor of Cyber Security and Director of the Institute of Cyber Security for Society (iCSS), University of Kent, U.K. His research interests are mostly about inter-disciplinary topics related to cyber security and privacy, human factors, digital forensics and cybercrime, multimedia computing, AI and data science. His work covers multiple application domains, including but not limited to cybercrime, social media analytics, digital health, smart cities, smart homes, and e-tourism. He received multiple awards and honors including the 2022 IEEE Transactions on Circuits and Systems Guillemin-Cauer Best Paper Award.
\end{IEEEbiography}

\begin{IEEEbiography}[{\includegraphics[width=1in,height=1.25in,clip,keepaspectratio]{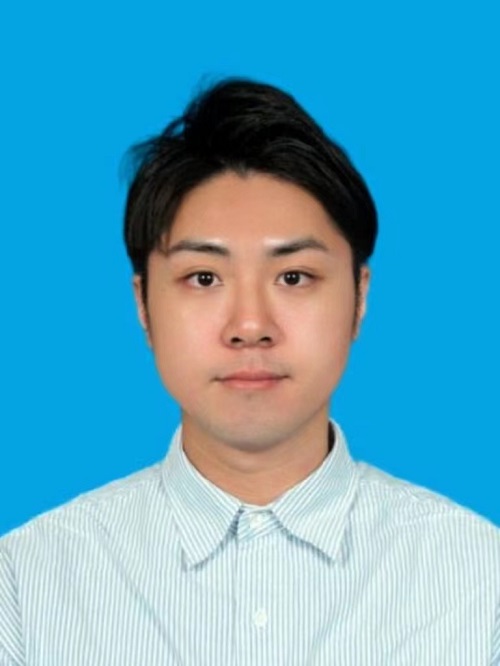}}]{Zhichao You} received his B.E.\ degree in Information and Computing Science from South China Agricultural University, Guangzhou, China in 2018, and his M.S.\ degree in Computer Science and Technology from Xidian University, China in 2021. He is now pursuing his Ph.D.\ degree at Xidian University. His research interests include federated learning and wireless network security.
\end{IEEEbiography}

\begin{IEEEbiography}[{\includegraphics[width=1in,height=1.25in,clip,keepaspectratio]{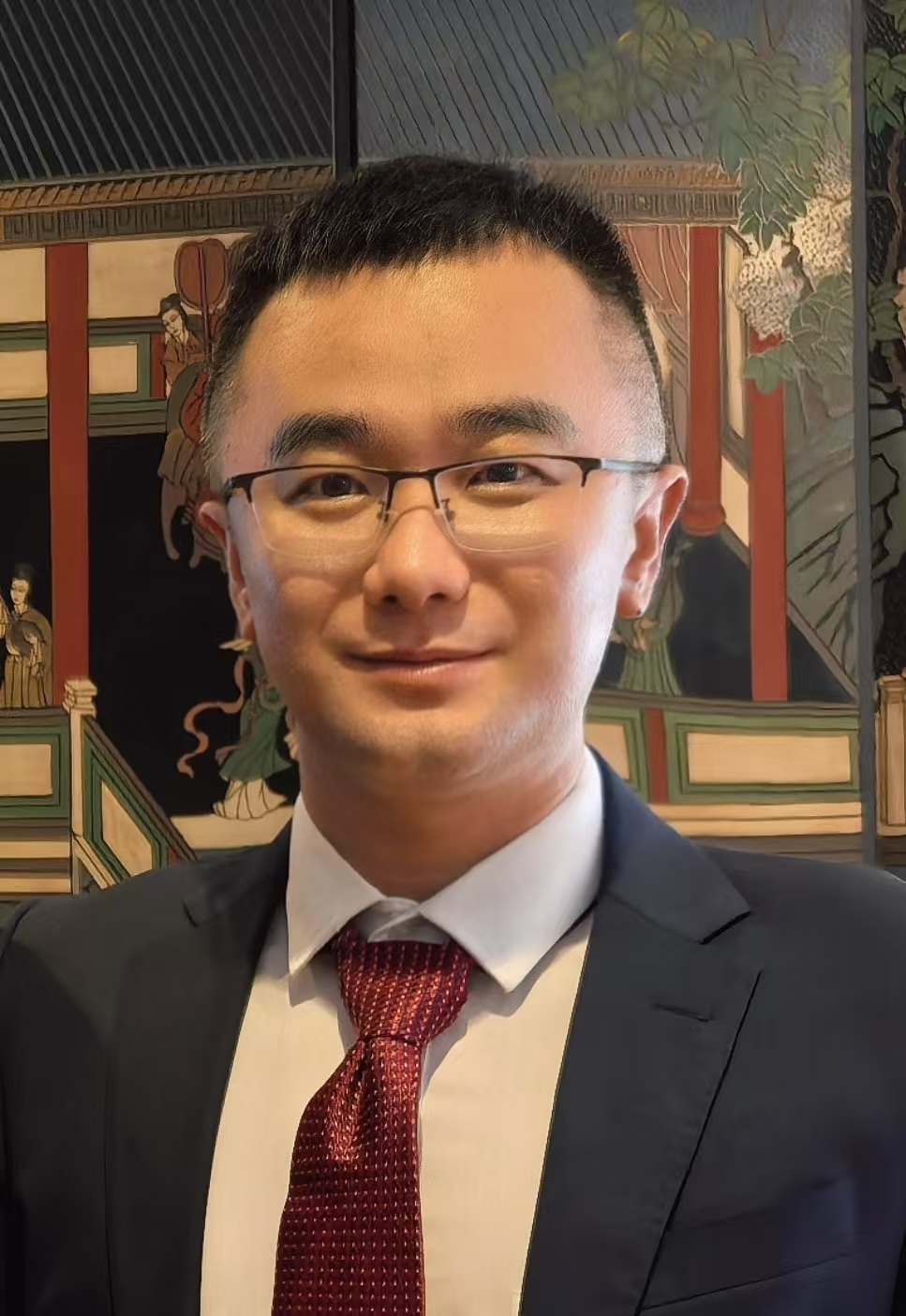}}]{Qiang Qu} received the PhD degree from Aarhus University, Aarhus, Denmark, supported by the GEOCrowd project under Marie Skodowska-Curie Actions. He is now a full professor with the Shenzhen Institutes of Advanced Technology, Chinese Academy of Sciences. His current research interests include large-scale data management and mining, blockchain and data intelligence.
\end{IEEEbiography}

\begin{IEEEbiography}[{\includegraphics[width=1in,height=1.25in,clip,keepaspectratio]{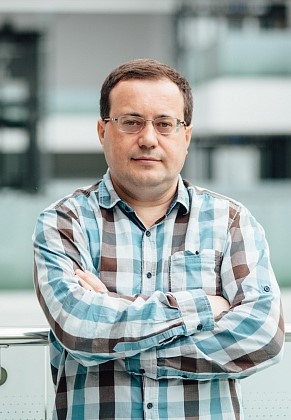}}]{Yaroslav Kholodov} had been a senior lecturer at the Department of Computational Mathematics of the Moscow Institute of Physics and Technology (MIPT) from 2021. In 2007, he became an Associate Professor. During 2009-2012 Yaroslav Kholodov had been a Chairman of the Organizing Board of the Annual MIPT School in High-Performance Computing for Young Scholars. During 2010-2012 Yaroslav had been the Head of the MIPT research and educational center ``High-Performance Computing and Distributed Computer Systems''. His key research area is represented by intelligent analysis of traffic data and road traffic modeling using adaptive control algorithms.
\end{IEEEbiography}

\begin{IEEEbiography}[{\includegraphics[width=1in,height=1.25in,clip,keepaspectratio]{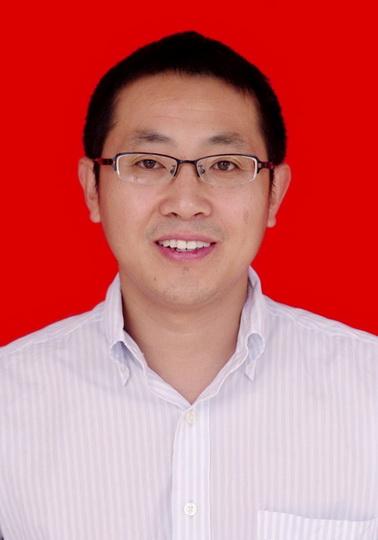}}]{Yulong Shen} received his B.S.\ and M.S.\ degrees in Computer Science and his Ph.D.\ degree in Cryptography from Xidian University, China, in 2002, 2005 and 2008, respectively. He is currently a Professor with the School of Computer Science and Technology, Xidian University, and also an Associate Director of the Shaanxi Key Laboratory of Network and System Security. He has also served on the technical program committees of several international conferences, including the NANA, the ICEBE, the INCoS, the CIS, and the SOWN. His research interests include wireless network security and cloud computing security.
\end{IEEEbiography}

\end{document}